\documentclass[a4paper,fleqn]{cas-dc}

\usepackage[numbers]{natbib}
\usepackage{textgreek}

\def\tsc#1{\csdef{#1}{\textsc{\lowercase{#1}}\xspace}}
\tsc{WGM}
\tsc{QE}

\begin{document}
\let\WriteBookmarks\relax
\def\floatpagepagefraction{1}
\def\textpagefraction{.001}

\shorttitle{}    

\shortauthors{Huang et al.}  

\title [mode = title]{Demographic-Aware Self-Supervised Anomaly Detection Pretraining for Equitable Rare Cardiac Diagnosis}  

\tnotemark[1] 


%

\author[1,2,3]{Chaoqin Huang}
\fnmark[*]
\credit{Developed the study, wrote the primary code, conducted the experiments, and generated the
figures, analyzed statistical data, drafted the initial manuscript}

\author[4]{Zi Zeng}
\fnmark[*]
\credit{Illustrated the clinical relevance of the study, provided essential hardware and data support, participated in data analysis, and contributed to drafting the manuscript}

\author[1,2]{Aofan Jiang}%
\fnmark[*]
\credit{Developed the study, wrote the primary code, conducted the experiments, and generated the
figures, analyzed statistical data, drafted the initial manuscript}

\author[4]{Yuchen Xu}
\credit{Revised the manuscript and engaged in in-depth discussions of the results}
\author[4]{Qing Cao}
\credit{Revised the manuscript and engaged in in-depth discussions of the results}
\author[4]{Kang Chen}
\credit{Revised the manuscript and engaged in in-depth discussions of the results}
\author[5]{Chenfei Chi}
\credit{Illustrated the clinical value, provided essential hardware and data support}
\author[1,2]{Yanfeng Wang}
\credit{critically revised the manuscript and engaged in in-depth discussions of the results, provided guide, supervision, and final approval for submission}
\author[1,2]{Ya Zhang}
\cormark[2]
\ead{ya_zhang@sjtu.edu.cn}
\credit{critically revised the manuscript and engaged in in-depth discussions of the results, provided guide, supervision, and final approval for submission}

\affiliation[1]{organization={Shanghai Jiao Tong University},
            city={Shanghai}, 
            country={China}}
\affiliation[2]{organization={Shanghai AI Laboratory},
            city={Shanghai},
            country={China}}
\affiliation[3]{organization={National University of Defense Technology},
            city={Wuhan},
            country={China}}
\affiliation[4]{organization={Ruijin Hospital, Shanghai Jiao Tong University School of Medicine},
            city={Shanghai},
            country={China}}
\affiliation[5]{organization={Renji Hospital, Shanghai Jiao Tong University School of Medicine},
            city={Shanghai},
            country={China}}


\tnotetext[1]{This work was funded by the National Key R\&D Program of China (No. 2022ZD0160702), STCSM (No. 22511106101, No. 18DZ2270700, No. 21DZ1100100), 111 plan (No. BP0719010), the Youth Science Fund of National Natural Science Foundation of China (No.7210040772).} 
\cortext[1]{These authors contributed equally and are co-first authors.}
\cortext[2]{Corresponding author at: Y. Zhang, Shanghai AI Laboratory, Shanghai Jiao Tong University, Shanghai, 200232, China}


\begin{abstract}
Rare cardiac anomalies are difficult to detect from electrocardiograms (ECGs) due to their long-tailed distribution with extremely limited case counts and demographic disparities in diagnostic performance. These limitations contribute to delayed recognition and uneven quality of care, creating an urgent need for a generalizable framework that enhances sensitivity while ensuring equity across diverse populations. In this study, we developed an AI-assisted two-stage ECG framework integrating self-supervised anomaly detection with demographic-aware representation learning. The first stage performs self-supervised anomaly detection pretraining by reconstructing masked global and local ECG signals, modeling signal trends, and predicting patient attributes to learn robust ECG representations without diagnostic labels. The pretrained model is then fine-tuned for multi-label ECG classification using asymmetric loss to better handle long-tail cardiac abnormalities, and additionally produces anomaly score maps for localization, with CPU-based optimization enabling practical deployment. Evaluated on a longitudinal cohort of over one million clinical ECGs, our method achieves an AUROC of 94.7\% for rare anomalies and reduces the common–rare performance gap by 73\%, while maintaining consistent diagnostic accuracy across age and sex groups. In conclusion, the proposed equity-aware AI framework demonstrates strong clinical utility, interpretable anomaly localization, and scalable performance across multiple cohorts, highlighting its potential to mitigate diagnostic disparities and advance equitable anomaly detection in biomedical signals and digital health. Source code is available at \url{https://github.com/MediaBrain-SJTU/Rare-ECG}.
\end{abstract}




\maketitle

\section{Introduction}
Cardiovascular diseases remain a leading cause of global mortality, with early and accurate diagnosis of cardiac anomalies through electrocardiograms (ECGs) playing a pivotal role in patient outcomes~\cite{ecgsurvey,wearble}. Despite advances in AI-driven ECG analysis, performance on rare anomalies remains a critical challenge due to their underrepresentation in long-tailed clinical datasets~\cite{lai2023practical, kalmady2024development, vaid2023foundational, nankani2022atrial}. While common conditions (\textit{e.g.}, sinus tachycardia, atrial fibrillation) dominate the clinical landscape, life-threatening rare anomalies (\textit{e.g.}, high-degree atrioventricular blocks, right ventricular infarction, sinus pause, high lateral wall myocardial infarction) are frequently missed~\cite{lai2023practical,wuclinical}. For example, second-degree type 2 atrioventricular block, present in fewer than 0.002\% of ECGs in our cohort, is often undetected by conventional models, delaying urgent interventions~\cite{takayaoutcomes}. Accurate detection of rare cardiac anomalies helps bridge diagnostic disparities, enabling timely treatment and improving patient outcomes. In addition to disease-specific disparities, demographic factors such as age and sex have been shown to modulate diagnostic accuracy~\cite{doi:10.1161/CIRCEP.119.007284, moss2010gender}, posing a further challenge to equitable AI deployment in cardiac care. Without explicit consideration of these factors, AI models risk amplifying existing health inequities.

However, achieving equitable detection across rare and common cardiac anomalies as well as different patient groups remains challenging under conventional supervised learning frameworks. Existing models primarily optimize for aggregate performance metrics on imbalanced datasets, often at the expense of underrepresented conditions. Although techniques such as contrastive learning~\cite{simclr} and synthetic data generation~\cite{goodfellow2020generative} have been explored to mitigate data imbalance, they frequently overlook the nuanced physiological variations (\textit{e.g.}, QRS and QT interval shifts) and demographic factors (\textit{e.g.}, age, sex) that influence ECG manifestations~\cite{doi:10.1161/CIRCEP.119.007284,rasmussenprinterval}. Moreover, most existing approaches focus narrowly on anomaly detection without providing precise localization or clinically interpretable outputs, limiting their utility in real-world decision-making~\cite{hughes2023deep, ukil2016iot}.

To overcome these limitations, we propose a two-stage diagnostic framework centered on fairness-aware representation learning. First, we introduce a self-supervised anomaly detection pretraining strategy tailored to ECG signals, where the model reconstructs masked segments at both the global (10-second rhythm) and local (heartbeat-level) scales. This process encourages the model to develop context-aware representations of normal cardiac patterns, thereby improving sensitivity to subtle deviations indicative of rare anomalies. Second, we incorporate demographic-aware modeling during pretraining by predicting patient-specific attributes such as age, sex, and physiological parameters. This auxiliary supervision guides the feature space toward demographic invariance, mitigating performance disparities across diverse patient subgroups. Together, these innovations aim to bridge the diagnostic equity gap between rare and common cardiac conditions while promoting consistent diagnostic sensitivity across demographic groups.

Our approach is rooted in the hypothesis that anomaly detection pretraining highlights clinically salient regions of interest, enabling the downstream classifier to focus on diagnostically meaningful features and precisely categorize anomaly types. While conventional anomaly detection is limited to binary discrimination~\cite{jiang2023multi,xu2022anomaly,zheng2022task}, our method achieves fine-grained categorization of rare anomalies, making it suitable for real-world clinical deployment. Unlike conventional contrastive~\cite{lai2023practical,simclr} or reconstruction-based methods~\cite{liu2022time,Tranad}, our framework explicitly aligns feature learning with fairness objectives, aiming to reduce both disease-type and demographic-driven diagnostic gaps. By integrating anomaly sensitivity and demographic robustness into feature representation, our approach further facilitates interpretability and subgroup generalization.

We validate our framework through multi-center trials spanning diverse geographic and clinical settings. The internal evaluation was performed on \textbf{ECG-LT}, one of the largest clinical ECG datasets to date, comprising 1,089,367 records with 116 cardiac types, 43 of which are rare anomalies absent in public datasets. Our model achieves 94.7\% AUROC, 92.2\% sensitivity, and 92.5\% specificity for rare anomalies, outperforming state-of-the-art methods by 8.9\% in AUROC and narrowing the performance gap with common conditions. Furthermore, the model’s anomaly localization capability, validated against cardiologist annotations with 76.5\% AUROC, provides interpretable insights by pinpointing critical regions directly aligned with clinical decision-making.

Externally, the model generalized robustly to the European PTB-XL dataset (+3.8\% AUROC) and Shanghai Renji Hospital’s clinical cohort (+8.1\% AUROC), addressing concerns about regional and institutional biases. Trials involving emergency department cardiologists demonstrated real-world clinical impact: AI-assisted cardiologists achieved 84.0\% diagnostic accuracy (vs. 77.3\% unaided) with a 32.5\% reduction in interpretation time, underscoring the framework’s potential to simultaneously enhance diagnostic precision and efficiency in high-stakes settings.

Our main contributions are summarized as follows:

\begin{itemize}
    \item \textbf{A novel two-stage diagnostic framework}: We integrate demographic-aware self-supervised anomaly detection pretraining to address the core challenges of long-tailed rare cardiac anomaly distribution and demographic-driven diagnostic disparities in ECG analysis.
    \item \textbf{A large-scale comprehensive benchmark}: We establish one of the largest and most diverse evaluation benchmarks for long-tailed ECG diagnosis, built on over one million clinically annotated ECG records spanning 116 cardiac anomaly types, including 43 rare conditions absent from existing public datasets.
    \item \textbf{Validated real-world clinical utility}: Simulated clinical validation in an emergency department setting demonstrates that our framework significantly improves cardiologists’ diagnostic accuracy (+6.7\%) and reduces ECG interpretation time (32.5\% reduction), supporting its practical deployment in time-sensitive acute cardiac care.
\end{itemize}

\section{Related Works}

In the realm of ECG analysis, current research has progressed along two primary avenues: computer-aided diagnosis and anomaly detection. The former, driven by advancements in deep learning, has emerged as the predominant approach. Conversely, the latter avenue, although less explored, presents novel opportunities for addressing long-tailed challenges in diagnosis. Our study represents the first systematic effort to integrate these two approaches, aiming to enhance ECG diagnosis particularly in cases with long-tailed distribution of anomalies.

\subsection{Computer-aided ECG diagnosis}

The advancement of deep learning has greatly enhanced the performance of computer-aided ECG diagnosis~\cite{liu2021deep}, enabling the field to transition from conventional rule-based diagnostic approaches~\cite{EBRAHIMI2020100033}. However, current research following the classification framework requires a large volume of labeled abnormal data~\cite{wagner2020ptb} and is limited to detecting only the specific anomaly types provided in the training data, such as arrhythmia classification~\cite{wang2023arrhythmia,rahul2022automatic}, atrial fibrillation classification~\cite{nankani2022atrial, feng2022novel}, or multi-label classification~\cite{du2021fm, cao2021practical}, thereby falling short of identifying unseen abnormal ECG signals. This constraint diminishes their practical utility in clinical diagnosis. Moreover, few studies have investigated the use of a pretraining–finetuning diagnostic framework on large-scale ECG datasets~\cite{lai2023practical,vaid2023foundational}. The limited research available suggests that while such approaches may yield strong diagnostic performance on common cases, their effectiveness on rare ECG conditions remains uncertain.

\subsection{Anomaly detection in electrocardiogram}

The scarcity and diversity of anomaly types, combined with the prevalence of normal ECGs, pose core challenges due to the limited availability of labeled anomalies for supervised training~\cite{ukil2016iot}. Consequently, anomaly detection tasks are typically framed as unsupervised learning tasks, relying on normal samples for training and identifying any samples that deviate from this norm as anomalies~\cite{pang2021deep}. The objective of ECG anomaly detection is to effectively distinguish between normal and anomalous samples, which can be viewed as a binary classification problem~\cite{chandola2009anomaly}. Current research frames ECG anomaly detection within the broader scope of time-series anomaly detection, primarily focusing on two key approaches: reconstruction-based methods and self-supervised learning-based methods~\cite{krishnan2022self}. Reconstruction-based methods~\cite{xu2022anomaly,zhou2019beatgan,Tranad} employ generative neural networks~\cite{li2019mad} to reconstruct normal samples, operating under the assumption that a model trained on normal samples will struggle to accurately reconstruct abnormal regions. By contrast, self-supervised learning-based methods~\cite{zheng2022task} leverage proxy tasks to improve representation learning of normal cardiac patterns.

\section{Methods}
\begin{figure*}
	\centering
	\includegraphics[width=2.0\columnwidth]{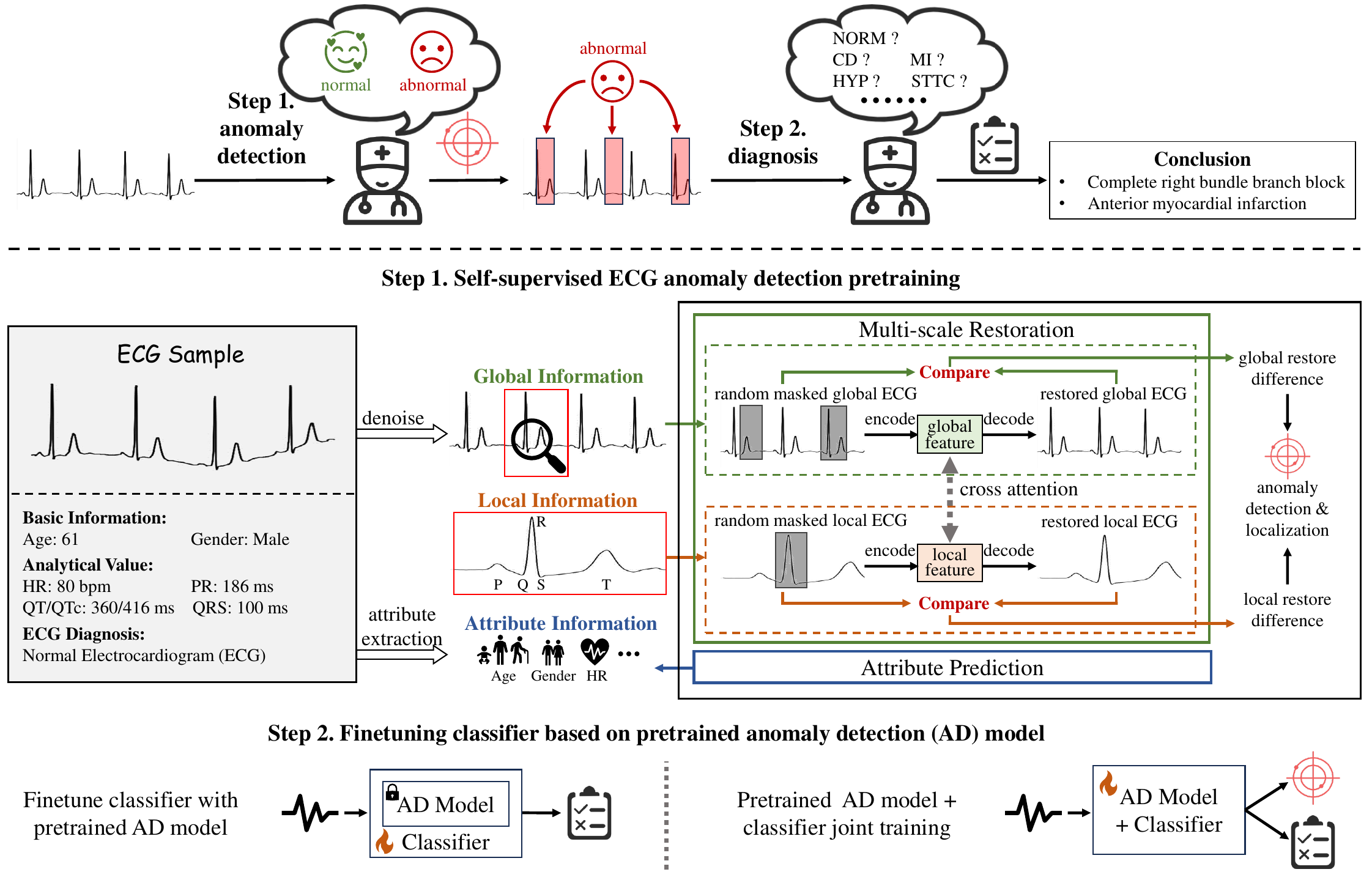}
	\caption{\textbf{The proposed two-stage ECG diagnosis framework.} Step 1: Self-supervised pretraining for ECG anomaly detection, training a model to detect abnormal patterns using global and local ECG features. Step 2: Fine-tuning the classifier based on the pretrained anomaly detection model for detailed diagnosis. This method enhances classification performance, particularly for less frequent cardiac conditions, by leveraging anomaly detection pretraining.}
	\label{fig:pipeline2stage}
\end{figure*}

This section outlines the learning framework (Fig.~\ref{fig:pipeline2stage}) that integrates anomaly detection pretraining into ECG classification, first providing a detailed overview of the proposed model, followed by its training procedure.

Each ECG training sample consists of two components: $\mathcal{S} = \{\mathcal{X}, \mathcal{T}\}$. Here, $\mathcal{X}$ denotes the signal components, comprising $D$ signal points extracted from the ECG sample as the framework’s input. $\mathcal{T}$ includes textual components, such as patient-specific data (e.g., age, gender), fundamental attributes (e.g., heart rate, PR interval), and diagnostic conclusions, which serve as supervision signals for the output. The framework aims to detect potential anomaly regions and predict cardiac anomaly classes based on the input ECG signal. Training involves two distinct phases: self-supervised anomaly detection and supervised classification.

\subsection{Self-supervised anomaly detection pretraining}
\begin{figure*}
	\centering
	\includegraphics[width=2.0\columnwidth]{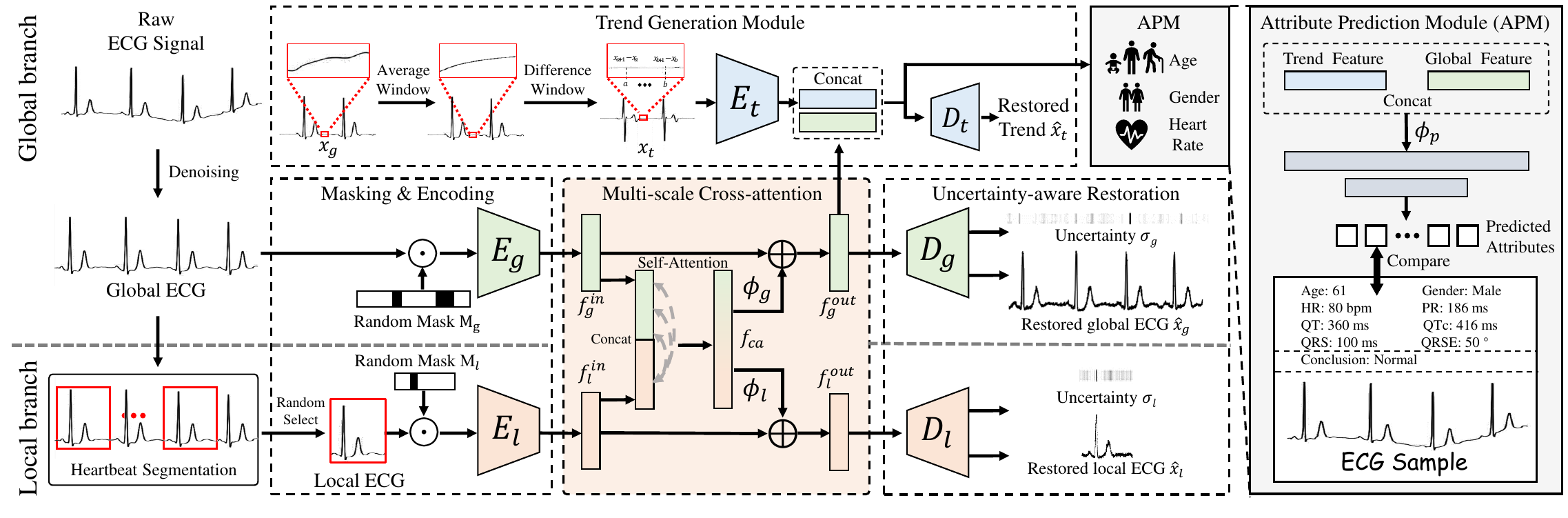}
	\caption{\textbf{The details of the multi-scale cross-restoration framework for ECG anomaly detection pretraining.}}
	\label{fig:pipeline_supp}
\end{figure*}

The self-supervised anomaly detection pretraining aims to restore randomly masked ECG signal segments and predict attribute data recorded within the samples. For a new test ECG, the model generates anomaly scores and detailed score maps by comparing the restored signal with the original. Notably, this process relies solely on attribute data (e.g., age, gender) as supervisory signals, eliminating the need for diagnostic information during training. As shown in Fig.~\ref{fig:pipeline_supp}, the framework, designed for detecting and localizing ECG anomalies, includes three key components: (i) multi-scale cross-restoration, (ii) trend-assisted restoration, and (iii) the attribute prediction module.

\subsubsection{Multi-scale cross-restoration}  
We start by analyzing the signal component $x$ in its global form $x_g \in \mathbb{R}^D$ and segmenting it into local heartbeats. Randomly selected heartbeats $x_l \in \mathbb{R}^d$ are paired with the global signal to form a global-local ECG pair for further processing. This pair is masked—scattered globally and continuously locally—before being processed by separate global and local encoders to extract pertinent features. Mimicking the diagnostic approach of expert cardiologists, who assess both overall ECG context and specific heartbeat details, our model employs a self-attention mechanism~\cite{vaswani2017attention} to integrate these features into a unified cross-attention feature set via concatenation. The self-attention mechanism dynamically weights these features, emphasizing the most relevant signal regions. Subsequently, two specialized decoders reconstruct the global signal $\hat{x}_g$ and local signal $\hat{x}_l$ from the integrated features.
This reconstruction process also produces restoration uncertainty maps $\sigma_g$ and $\sigma_l$, which visualize the challenges encountered during signal restoration, offering insights into task complexity. To improve restoration performance, we adopt an uncertainty-aware restoration loss~\cite{mao2020uncertainty}:
\begin{equation}
\mathcal{L}_{res} = \sum_{k=1}^D [\frac{(x_g^k - \hat{x}_g^k)^2}{\sigma_g^k} + \log \sigma_g^k]+\sum_{k=1}^d [\frac{(x_l^k - \hat{x}_l^k)^2}{\sigma_l^k} + \log \sigma_l^k],
\end{equation}
where each term of the loss function normalizes the squared error by its uncertainty (first term) and includes a logarithmic penalty (second term) to discourage excessive uncertainty across all signal points. Here, the superscript $k$ denotes the $k$-th element’s position within the signal.

\subsubsection{Trend-assisted restoration}  
To improve the model's ability to restore and interpret global ECG signals, we generate a temporal trend signal $x_t \in \mathbb{R}^D$ from the global signal $x_g$. This trend is derived by smoothing $x_g$ with an averaging window during convolution, followed by a difference window to capture local variations. An auxiliary autoencoder then reconstructs the original signal by leveraging both trend and global
features. The reconstructed signal $\hat{x}_t$ is optimized via Euclidean distance: 
\begin{equation}
\mathcal{L}_{trend} = \sum_{k=1}^D (x_g^k - \hat{x}_t^k)^2. 
\end{equation}
Notably, this combination of global features and trend features encapsulates critical ECG signal characteristics, supporting subsequent attribute prediction and classification tasks.

\subsubsection{Attribute prediction module}  
We leverage the textual component $\mathcal{T}$, specifically extracting its attribute subset  $\mathcal{T}_{attr}$ (e.g., patient age, gender, heart rate, PR interval). These attributes are normalized into a vector $t_{attr} := [t_1, \dots, t_m]$ and used as supervisory signals. Leveraging the feature combination from the previous section, we predict these attributes as $\hat{t}_{attr} := [\hat{t}_1, \dots, \hat{t}_m]$ using an auxiliary multi-layer perceptron. The prediction accuracy is optimized via mean squared error loss: 
\begin{equation}
\mathcal{L}_{pred}(t_{attr}, \hat{t}_{attr}) = \frac{1}{m} \sum_{i=1}^m (t_i - \hat{t}_i)^2,
\end{equation}
improving the model’s ability to capture the influence of these attributes on ECG diagnosis.

The anomaly detection framework is optimized with a composite loss function:
\begin{equation}
\mathcal{L}_{AD} = \mathcal{L}_{res} + \alpha \mathcal{L}_{trend} + \beta  \mathcal{L}_{pred},
\end{equation}
where $\alpha$ and $\beta$ are trade-off parameters balancing the contributions of each loss term. For simplicity, we set $\alpha = \beta = 1.0$. This framework effectively identifies and localizes anomalies in a binary classification setting, labeling ECGs deviating from the norm as anomalies. A complementary classification network further provides precise ECG diagnostic outcomes for clinical use.

During testing, an ECG instance $x_{test}$ is processed by iteratively selecting local ECG segments, ${x_{l,m}, m = 1,\dots,M}$ from segmented heartbeats, contrasting with the random selection used in training. These local segments are paired with the global test ECG to form an input pair for the framework, which generates an anomaly score for detection, a score map for localization, and a classification result for clinical diagnosis. 
The score map $S(x_{test})$ combines predictions from the global and local branches, expressed as $S(x_{test}) = S_g(x_{test}) + S_l(x_{test})$, where $S_g(x_{test})$ and $S_l(x_{test})$ represent the score maps for the global and local branches, respectively. The global branch score is defined as: 
\begin{equation}
S_g(x_{test}) = \frac{(x_{test} - \hat{x}_{test})^2}{\sigma_g} + (x_{test} - \hat{x}_t)^2.
\end{equation}
The local score map aggregates contributions from individual heartbeat segments, formulated as:
\begin{equation}
S_l(x_{test}) = \sum_{m=1}^M I_m \odot \frac{(x_{l,m} - \hat{x}_{l,m})^2}{\sigma_{l,m}},
\end{equation}
where $I_m \in \mathbb{R}^D$ is a binary indicator vector, taking the value of one only at the position of the $m$-th segmented heartbeat. The anomaly score $A(x_{test})$ is the mean value of all signal points in $S(x_{test})$. The classification result $Y(x_{test})$ is a fixed-length vector, where each element corresponds to a specific anomaly type, representing its probability. Higher values in the score map, anomaly score, and classification result indicate a greater
likelihood of anomalies.

\subsection{Supervised classification}
\label{sec:claarch}
Following self-supervised anomaly detection pretraining, the model is designed to perform multi-label classification of ECG signals, accommodating the presence of multiple anomalies within a single recording. The pretrained anomaly detection model acts as a feature extractor for ECG signals, which can be employed in two ways: (i) a two-stage process where a separate classifier is trained on the extracted features, or (ii) an end-to-end joint training of the anomaly detection model and classifier.
Diagnostic annotations are encoded as a multi-hot vector $y = [y_1, \dots,y_n] \in \{0,1\}^n$, 
where $n$ represents the number of anomaly ECG types. This label serves as the ground-truth supervision signal, indicating the presence or absence of each ECG type. An additional classification head, built upon the same combined features used for attribute prediction, outputs probability predictions $\hat{y} = [\hat{y}_1, \dots, \hat{y}_n]$ for potential cardiac conditions. To optimize these predictions against the ground-truth labels, we adopt the Asymmetric Loss function~\cite{benbaruch2020asymmetric}, defined as:
\begin{equation}
\mathcal{L}_{cls} = 
    \sum_{k=1}^n [-{y}_k (1 - \hat{y}_k)^{\gamma_+} \log(\hat{y}_k) - (1 - y_k) ({\hat{y}_{k}^m})^{\gamma_-} \log(1 - {\hat{y}_{k}^m})],
\end{equation}
where $\hat{y}_{k}^m = \max(\hat{y}_{k} - \tau, 0)$ applies a hard threshold ($\tau$ is a fixed margin) to down-weight easily classified negatives, excluding them when their probability is low, thereby directing the model to prioritize challenging anomaly cases. $\gamma_+$ and $\gamma_-$ are positive and negative focusing weight parameters, respectively, controlling the influence of positive and negative labels. Given our emphasis on positive labels (indicating present anomalies), $\gamma_-$ is set greater than $\gamma_+$, enhancing the model’s focus on rare anomaly features.

Our two-stage diagnostic framework integrates self-supervised anomaly detection with supervised classification through fine-tuning. We explore two strategies. First, the pretrained anomaly detection model serves as a fixed feature extractor, where global and trend features from the encoder are concatenated to form diagnostic representations. The classifier is trained on these features using the loss function $\mathcal{L}_{cls}$. Alternatively, we propose a joint training strategy, simultaneously optimizing the anomaly detection model and classifier. Here, the ECG signal is processed by the anomaly detection model, with extracted features fed into the classifier. The joint objective combines both losses as $\mathcal{L}_{AD} + \mathcal{L}_{cls}$, enabling end-to-end optimization of the framework.

\section{Experiments}
\subsection{Datasets}

\subsubsection{ECG-LT: internal dataset}
\label{sec:dataruijin}
\begin{figure*}
	\centering
	\includegraphics[width=1.8\columnwidth]{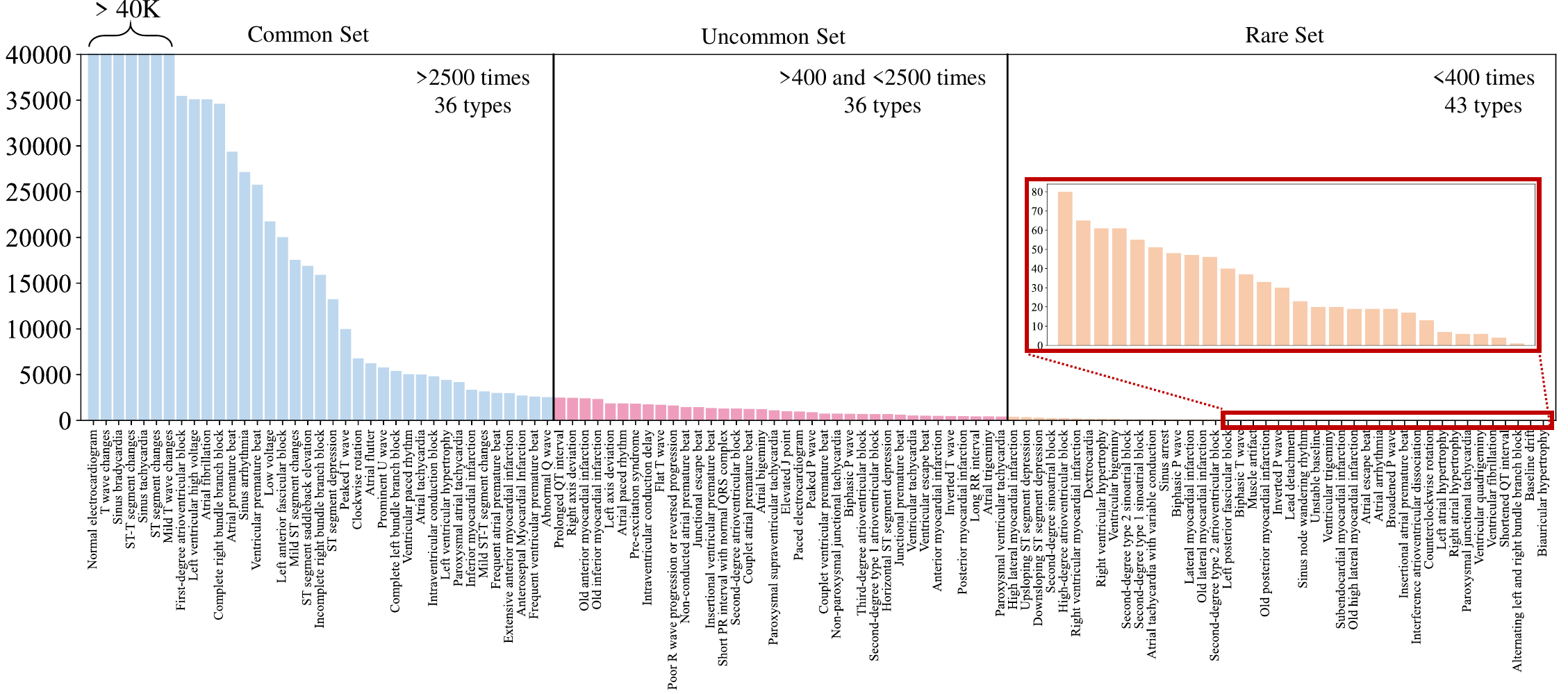}
	\caption{\textbf{The long-tailed (extremely imbalanced) distribution of cardiac disease types across the ECG-LT dataset,} with cardiac disease types divided into common set, uncommon set, and rare set based on their frequency of occurrence. The red box highlights the expanded view of part of the Rare Set, where cardiac disease types occur fewer than 80 times.}
	\label{fig:datamethod}
\end{figure*}

We evaluated models on a comprehensive real-world clinical dataset, consisting of \textbf{1,089,367 ECG samples} collected between 2012 and 2021. Each sample includes an ECG signal and a diagnostic summary identifying specific anomalies. The dataset spans 116 unique cardiac anomaly types. Figure~\ref{fig:datamethod} illustrates the pronounced long-tailed distribution: 43 anomaly types occur fewer than 400 times, many of which are absent from public ECG datasets due to their low prevalence, highlighting the dataset’s unique clinical value.\footnote{For example, ECG-LT includes 23 cases of dextrocardia—a congenital anomaly where the heart is positioned on the right side of the chest—which is often misdiagnosed as myocardial infarction or conduction abnormalities. It also contains cases of alternating left and right bundle branch block (ALRBBB), a rare conduction disorder with intermittent shifts between left and right bundle branch block, linked to increased risks of heart failure, arrhythmias, and sudden cardiac death, potentially indicating impending complete heart block.}

\begin{figure*}
	\centering
	\includegraphics[width=2.0\columnwidth]{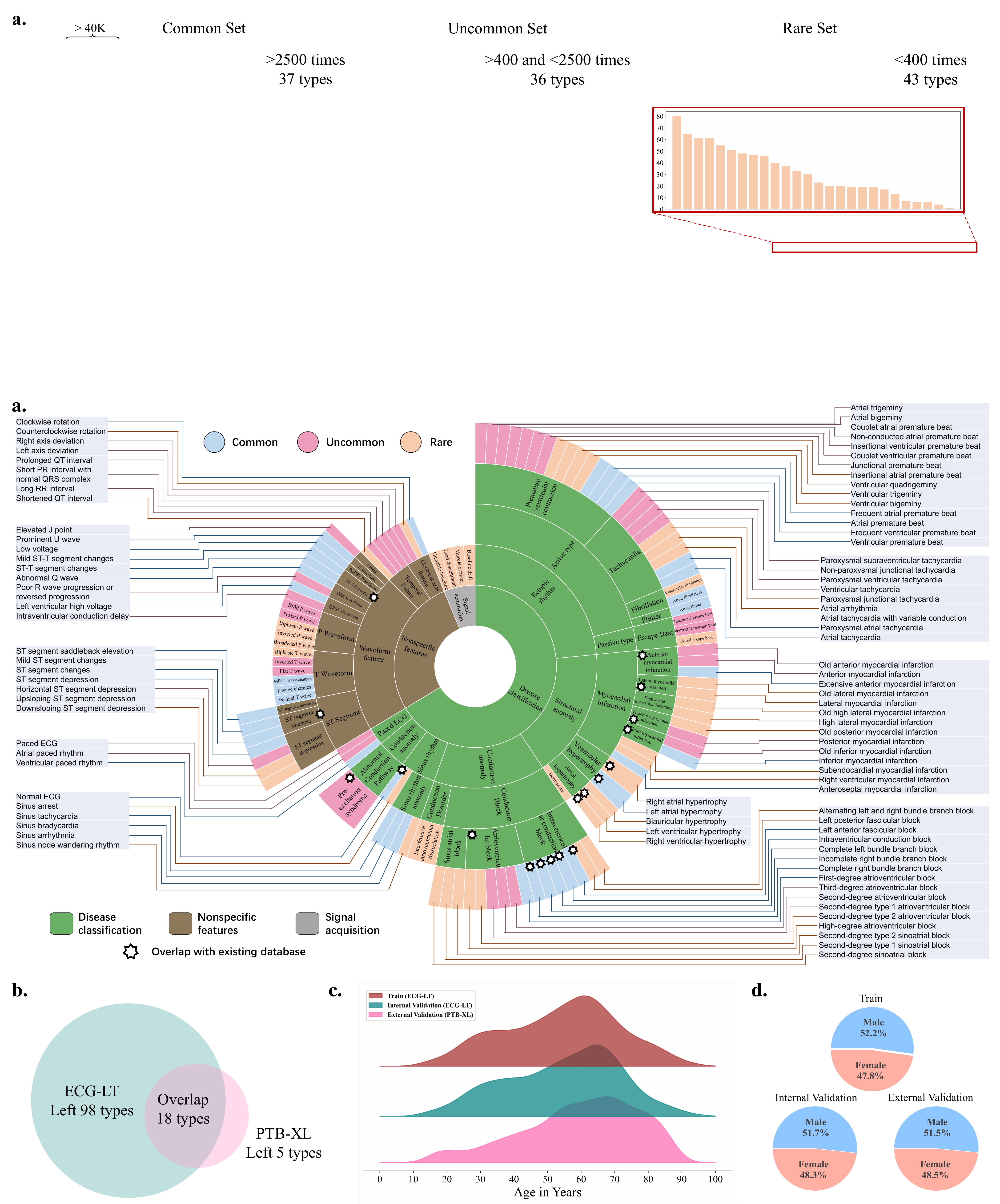}
	\caption{\textbf{Comprehensive analysis of the novel ECG-LT dataset.} a. Hierarchical architecture of cardiac types. b.
Comparison of the number of cardiac types in the ECG-LT dataset to those in existing ECG databases. c. Age distribution across
the training, internal validation, and external validation sets. d. Gender distribution across the training, internal validation, and
external validation sets.}
	\label{fig:data_type}
\end{figure*}

All anomaly types were manually diagnosed by experienced cardiologists over nine years, covering clinical conditions, signal features, and acquisition artifacts. The hierarchical structure of these anomaly types is detailed in Fig.~\ref{fig:data_type}a.The dataset also includes patient demographics (e.g., gender, age) and physiological parameters (e.g., heart rate, PR, QT, corrected QT, QRS intervals). ECG samples were converted into 12-lead time-series data, with each lead capturing 2.5 seconds (except lead II, which captured 10 seconds) at a 500 Hz sampling rate.

For training, we used ECGs collected up to 2020, totaling 416,951 normal and 482,976 abnormal samples. Internal validation was performed on 2021 ECGs, comprising 94,304 normal and 95,136 abnormal samples, ensuring no overlap with the training set. Validation ECGs were stratified into three test sets: common (37 types, $>2,500$ occurrences), uncommon (36 types), and rare (43 types, $<400$ occurrences). Age and gender distributions were consistent across training and validation sets (average age: 54), as shown in Figures~\ref{fig:data_type}c and \ref{fig:data_type}d. Additionally, two experienced cardiologists manually annotated anomaly locations at the signal-point level on 500 ECGs to create a test set for precise localization evaluation.

\subsubsection{External datasets}
\label{sec:dataopen}

The open-source PTB-XL dataset~\cite{wagner2020ptb} comprises 21,837 clinical 12-lead ECGs from 18,885 patients, each lasting 10 seconds and sampled at 500 Hz. Collected between October 1989 and June 1996 using Schiller AG devices, the dataset includes 23 distinct ECG subclasses, 18 of which overlap with our internal ECG-LT dataset (see Figures~\ref{fig:data_type}a and \ref{fig:data_type}b). The five ECG types unique to the PTB-XL dataset arise from specific pathological delineations in diagnostic conclusions, a protocol not applied in ECG-LT due to differing clinical practices. For external validation, we used a split test set of 2,198 ECGs, following the PTB-XL framework. The age and gender distributions between internal and external validation sets are compared in Fig.~\ref{fig:data_type}c and \ref{fig:data_type}d, showing 51.5\% male representation in the external set (vs. 51.7\% in the internal set) and a higher proportion of patients aged 70–90 years. The average age in the external validation set is 59, five years older than the internal dataset’s average.

The clinical Renji dataset comprises 1,385 ECG recordings with manual diagnostic conclusions, collected at Shanghai Renji Hospital in 2024. Unlike PTB-XL, the Renji dataset exclusively includes atypical cases rarely seen in routine practice, curated with cardiologist input to ensure uniqueness. Used solely for external validation, this dataset originates from a distinct hospital setting, ensuring no overlap with the training or internal validation sets. It encompasses 26 distinct anomaly types, detailed in Supplementary Table~S4.

ECG signals were preprocessed using a Butterworth filter and Notch filter~\cite{van2019heartpy} to remove high-frequency noise and mitigate baseline wander. Multiple ECG lead signals were concatenated in the clinical report order, with 2.5-second leads combined into a 10-second segment and lead II retained as a single channel, yielding a four-channel input signal. R-peaks were detected using an adaptive threshold method~\cite{van2019rpeak}, which avoids learnable parameters. These R-peak positions segment the ECG into individual heartbeats. The global ECG signal comprises 5000 signal points, while each local heartbeat segment contains 500 points, i.e., D = 5000, d = 500.

\subsection{Evaluation Metrics}
We evaluate ECG classification performance using the area under the receiver operating characteristic curve (AUROC), sensitivity, and specificity, with $p$-values computed via McNemar's test to compare baseline predictions with our approach. For anomaly detection, we report the F1 score and precision at a fixed 90\% recall. For anomaly localization, we utilize the Dice coefficient to measure the overlap between predictions and expert annotations.

\subsection{Implementation details}

All experiments are implemented using the PyTorch deep learning framework. The anomaly detection module employs a convolutional autoencoder architecture, while the classification network adopts a 1D-ResNet model. Training uses the AdamW optimizer with an initial learning rate of $1 \times 10^{-4}$ and a weight decay of $1 \times 10^{-5}$. The models are trained for 50 epochs with a batch size of 32 and a cosine learning rate schedule.

\subsection{Comparative analysis}To validate the effectiveness of the proposed framework, we compared it with several representative baseline methods.We evaluated a model trained from scratch, which directly performs ECG classification without pretraining, to examine the contribution of anomaly detection pretraining.We then compared our method with three representative self-supervised and anomaly detection pretraining approaches, including Contrastive Learning, Momentum Contrast (MoCo~\cite{lai2023practical}), and Task-Specific Learning (TSL~\cite{zheng2022task}).

For fair comparisons, all methods used the same downstream classifier and were evaluated under identical training and testing settings. The pretrained representations were either used as frozen feature extractors with classifier fine-tuning or through joint training with the classifier.

\subsection{Simulated clinical validation}
To evaluate real-world applicability, we conducted a validation study in the emergency department of Ruijin Hospital from May to July 2024. The study included 238 ECG recordings covering 50 cardiac anomaly types. The proposed model was deployed without additional fine-tuning and used as a decision-support tool.

Each ECG was evaluated under three diagnostic settings: (i) rapid diagnosis, where cardiologists interpreted ECGs under strict time constraints to simulate emergency conditions; (ii) independent diagnosis, where cardiologists performed routine interpretation without time limits; and (iii) AI-assisted diagnosis, where cardiologists were provided with the top five predictions generated by the model. All cardiologists had comparable clinical experience, and the reference diagnosis was determined by an independent senior cardiologist.

\section{Results}
\subsection{Evaluation across internal and external cohorts}

\begin{table*}[t]
 \caption{
 Performance of ECG diagnosis across multiple evaluation datasets, including internal validation on ECG-LT and external validation on PTB-XL and Renji. The best-performing method is shown in \textbf{bold}, and the second-best is \underline{underlined}. }
\centering
\scalebox{0.9}{
\begin{tabular}{cccccccccc}\toprule
& & & & & \multicolumn{5}{c}{Pretraining + Finetuning}\\
\Xcline{6-10}{0.3pt}
Dataset & \makecell{Evaluation\\ Data} & \makecell{ECG\\Sample\\Count} & Metrics & \makecell{Train from\\Scratch} & \makecell{Contrastive\\Learning} & 
MoCo &
\makecell{AD:TSL} & \makecell{AD:Ours\\Finetuning} & \makecell{AD:Ours\\Joint-train}\\
\hline
\multirow{12}{*}{\makecell{Internal\\ Validation\\ (ECG-LT)}} & \multirow{3}{*}{\makecell{All\\ Test Data}} & \multirow{3}{*}{189,440} & AUROC & 0.910$_{\pm0.053}$ & 0.923$_{\pm0.050}$ & 0.906$_{\pm0.054}$ & 0.925$_{\pm0.049}$  & \underline{0.948$_{\pm0.041}$} & \textbf{0.964$_{\pm0.035}$}\\
& & & Sensitivity & 0.879$_{\pm0.061}^{*}$ &  0.882$_{\pm0.060}^{*}$ & 0.844$_{\pm0.068}^{*}$ & 0.878$_{\pm0.061}^{*}$ & \underline{0.925$_{\pm0.049}^{*}$} & \textbf{0.931$_{\pm0.047}$}\\
& & & Specificity & 0.830$_{\pm0.070}^{*}$ & 0.888$_{\pm0.059}^{*}$ & 0.876$_{\pm0.062}^{*}$ & 0.879$_{\pm0.061}^{*}$ & \underline{0.889$_{\pm0.059}^{*}$} & \textbf{0.925$_{\pm0.049}$}\\
\Xcline{2-10}{0.3pt}
& \multirow{3}{*}{\makecell{Common\\ Test Data}} & \multirow{3}{*}{183,810} & AUROC & 0.940$_{\pm0.077}$ & 0.937$_{\pm0.078}$ & \underline{0.964$_{\pm0.060}$} & 0.932$_{\pm0.081}$ & 0.962$_{\pm0.062}$ & \textbf{0.969$_{\pm0.056}$}\\
& & & Sensitivity & 0.887$_{\pm0.102}^{*}$ & 0.886$_{\pm0.102}^{*}$ & \underline{0.920$_{\pm0.087}^{*}$} & 0.890$_{\pm0.100}^{*}$ & 0.919$_{\pm0.088}^{*}$ & \textbf{0.935$_{\pm0.079}$}\\
& & & Specificity & 0.864$_{\pm0.110}^{*}$ & 0.880$_{\pm0.105}^{*}$ & \textbf{0.919$_{\pm0.088}^{*}$} & 0.856$_{\pm0.113}^{*}$ & 0.905$_{\pm0.094}^{*}$ & \underline{0.913$_{\pm0.091}$}\\
\Xcline{2-10}{0.3pt}
& \multirow{3}{*}{\makecell{Uncommon\\ Test Data}} & \multirow{3}{*}{5,253} & AUROC & 0.932$_{\pm0.082}$ & 0.948$_{\pm0.073}$ & 0.947$_{\pm0.073}$ & 0.950$_{\pm0.071}$ & \underline{0.968$_{\pm0.057}$} & \textbf{0.977$_{\pm0.049}$}\\
& & & Sensitivity & 0.872$_{\pm0.109}^{*}$ & 0.896$_{\pm0.099}^{*}$ & 0.894$_{\pm0.101}^{*}$ & 0.896$_{\pm0.099}^{*}$ & \underline{0.915$_{\pm0.091}^{*}$} & \textbf{0.937$_{\pm0.079}$}\\
& & & Specificity & 0.865$_{\pm0.111}^{*}$ & 0.897$_{\pm0.099}^{*}$ & 0.911$_{\pm0.093}^{*}$ &  0.899$_{\pm0.098}^{*}$ & \underline{0.921$_{\pm0.088}^{*}$} & \textbf{0.938$_{\pm0.079}$}\\
\Xcline{2-10}{0.3pt}
& \multirow{3}{*}{\makecell{Rare\\ Test Data}} & \multirow{3}{*}{386} & AUROC & 0.858$_{\pm0.112}$ & 0.886$_{\pm0.102}$ & 0.808$_{\pm0.127}$ & 0.893$_{\pm0.099}$ & \underline{0.914$_{\pm0.090}$} & \textbf{0.947$_{\pm0.072}$}\\
& & & Sensitivity & 0.876$_{\pm0.106}^{*}$ & 0.865$_{\pm0.110}^{*}$ & 0.721$_{\pm0.145}^{*}$ & 0.848$_{\pm0.116}^{*}$ & \underline{0.877$_{\pm0.106}^{*}$} & \textbf{0.922$_{\pm0.086}$}\\
& & & Specificity & 0.762$_{\pm0.137}^{*}$ & 0.886$_{\pm0.102}^{*}$ & 0.801$_{\pm0.129}^{*}$ & 0.882$_{\pm0.104}^{*}$ & \underline{0.912$_{\pm0.091}^{*}$} & \textbf{0.925$_{\pm0.085}$}\\
\hline
\multirow{3}{*}{\makecell{External\\ Validation\\ (PTB-XL)}} & \multirow{3}{*}{\makecell{Common\\ Test Data}} & \multirow{3}{*}{2,198} & AUROC & 0.858$_{\pm0.143}$ & 0.859$_{\pm0.142}$ & \underline{0.878$_{\pm0.134}$} & 0.864$_{\pm0.140}$ & 0.876$_{\pm0.135}$ & \textbf{0.896$_{\pm0.125}$}\\
& & & Sensitivity & 0.774$_{\pm0.171}^{*}$ & 0.814$_{\pm0.159}^{*}$ & 0.813$_{\pm0.159}^{*}$ & 0.797$_{\pm0.164}^{*}$ & \underline{0.818$_{\pm0.158}^{*}$} & \textbf{0.835$_{\pm0.152}$}\\
& & & Specificity & 0.790$_{\pm0.166}^{*}$ & 0.776$_{\pm0.170}^{*}$ & \underline{0.826$_{\pm0.155}^{*}$} & 0.794$_{\pm0.165}^{*}$ & 0.805$_{\pm0.162}^{*}$ & \textbf{0.848$_{\pm0.147}$}\\
\hline
\multirow{3}{*}{\makecell{External\\ Validation\\ (Renji)}} & \multirow{3}{*}{\makecell{Long-tail\\ Test Data}} & \multirow{3}{*}{1,385} & AUROC & 0.773$_{\pm0.150}$ & 0.779$_{\pm0.148}$ & 0.788$_{\pm0.167}$ & 0.768$_{\pm0.151}$ & \underline{0.790$_{\pm0.146}$} & \textbf{0.854$_{\pm0.128}$}\\
& & & Sensitivity & 0.726$_{\pm0.160}^{*}$ & 0.760$_{\pm0.153}^{*}$ & 0.753$_{\pm0.176}^{*}$ & 0.722$_{\pm0.160}^{*}$ & \underline{0.763$_{\pm0.152}^{*}$} & \textbf{0.842$_{\pm0.130}$}\\
& & & Specificity & \underline{0.820$_{\pm0.137}^{*}$} & 0.797$_{\pm0.144}^{*}$ & 0.813$_{\pm0.159}^{*}$ & 0.814$_{\pm0.139}^{*}$ & 0.818$_{\pm0.138}^{*}$ & \textbf{0.866$_{\pm0.122}$}\\
\bottomrule
\multicolumn{4}{l}{\small AD: Anomaly Detection.}\\
\multicolumn{10}{l}{\small $^{**}$: Denotes a \textit{p}-value $< 0.0001$ compared with the proposed Joint-train method.}
\end{tabular}
}
\label{tab:class_ruijin}
\end{table*}

To comprehensively assess the diagnostic performance and generalizability of the framework, we conduct rigorous validation on both internal and external datasets, with a particular focus on its effectiveness in handling long-tailed distributions.

We compare the proposed approach against a model trained from scratch~\cite{he2016deep} (i.e., without pretraining) as well as three distinct pretraining strategies: (i) Task-Specific Learning (TSL)~\cite{zheng2022task}, a state-of-the-art anomaly detection approach; (ii) contrastive learning~\cite{simclr}, a widely adopted self-supervised technique that enhances representation learning by grouping similar data points and separating dissimilar ones in the latent space; and (iii) Momentum Contrast (MoCo)~\cite{lai2023practical}, a pretraining strategy optimized for electrocardiogram (ECG) signals.

\subsubsection{Internal validation (ECG-LT Dataset)}
We conduct internal validation on the ECG-LT dataset, comprising 1,089,367 records, with the test data stratified into common (183,810 samples), uncommon (5,253 samples), and rare (386 samples) subsets (Fig.~\ref{fig:Diagnosis}a). As shown in Table~\ref{tab:class_ruijin}, the model trained from scratch exhibits a notable performance gap, with an 8.2\% AUROC drop between common (94.0\%) and rare (85.8\%) subsets. In contrast, our pretrained anomaly detection model, initially trained on normal ECG data, significantly mitigates this disparity. When used as a frozen feature extractor (AD: Finetuning), it improves the rare-class AUROC to 91.4\%. Joint training of anomaly detection and classification modules (AD: Joint-train) further enhances performance, achieving 94.7\% AUROC, 92.2\% sensitivity, and 92.5\% specificity for rare classes, and 96.4\% AUROC, 93.1\% sensitivity, and 92.5\% specificity across all test data, reducing the common-rare AUROC gap to 2.2\%. Notably, most of the rarest conditions in the subset achieve an AUROC above 0.95 (Fig.~\ref{fig:Diagnosis}a), underscoring the effectiveness of anomaly detection pretraining in tackling long-tailed classification challenges. Hereafter, “our method” refers to this joint-training approach, which is used in all subsequent validations.

Comparative analysis revealed critical limitations in alternative pretraining approaches. Compared to our method, contrastive learning underperformed by 4.1\% in AUROC, 4.9\% in sensitivity and 3.7\% in specificity on rare classes ($p$-value $< 0.0001$), despite its strength in learning invariant ECG representations through data augmentation. MoCo, optimized for ECG pretraining, achieved 96.4\% AUROC on common cases but dropped to 80.8\% AUROC on rare classes, a 13.9\% gap compared to our method ($p$-value $< 0.0001$), revealing its poor generalization to low-incidence conditions. TSL, another anomaly detection pretraining approach, showed a smaller common-to-rare AUROC gap (93.2\% vs. 89.3\%) but still lagged behind our framework by 5.4\% in rare-class AUROC ($p$-value $< 0.0001$). These results collectively demonstrate the superiority of our pretraining strategy in achieving balanced diagnostic accuracy across all anomaly frequencies,including rare ECG diagnoses.

\subsubsection{External validation}
We evaluate the generalizability of our model using two independent external cohorts: the PTB-XL Benchmark and the Renji Clinical Cohort.

The PTB-XL dataset differs from the proprietary ECG-LT dataset in age distribution, signal quality, and ECG signal types (Fig.~\ref{fig:data_type}). Under linear probing (training only the final classifier layer), our model achieves an AUROC of 89.6\%, sensitivity of 83.5\%, and specificity of 84.8\%. It outperforms contrastive learning, MoCo, and TSL pretraining by 3.7\% ($p$-value $< 0.0001$), 1.8\% ($p$-value $< 0.0001$), and 3.2\% ($p$-value $< 0.0001$) in AUROC, respectively (Table~\ref{tab:class_ruijin}). This demonstrates the model’s robustness across diverse ECG characteristics despite limited retraining.

On 1,385 long-tail ECGs from  Renji Clinical Cohort, without fine-tuning, our model achieved an AUROC of 85.4\%, sensitivity of 84.2\%, and specificity of 86.6\%. It surpassed baseline methods by 6.4\% ($p$-value $< 0.0001$) in AUROC, 7.9\% ($p$-value $< 0.0001$) in  sensitivity and 4.6\% ($p$-value $< 0.0001$) in specificity (Table~\ref{tab:class_ruijin}). These results highlight the model’s superior generalization to long-tailed distributions in real-world clinical settings.

\subsection{Diagnosis fairness}
Ensuring consistent diagnostic performance across key demographic subgroups is critical for clinical applicability. To assess fairness, we evaluate model accuracy stratified by sex and age. Specifically, we compare AUROC, sensitivity, and specificity for male versus female participants, and for different age ranges spanning from pediatric to geriatric populations.Results indicate that males and females exhibit comparable diagnostic outcomes, as illustrated in Fig.~\ref{fig:Diagnosis}b. Male patients show a modest improvement over female patients, with an increase of 0.8\%, 1.2\%, and 1.7\% in AUROC, sensitivity, and specificity, respectively. Regarding age stratification (Fig.~\ref{fig:Diagnosis}c), the model performs slightly less effectively in individuals under 10 years old (AUROC of 88\% and specificity of 78\%) and in those over 90 years old (specificity of 85\%). In contrast, the age groups ranging from 10 to 90 years maintain similar diagnostic metrics (AUROCs and specificities above 90\%). These findings demonstrate that while the model retains robust performance across most demographic groups, further refinements or tailored calibration may be beneficial for populations at the extremes of age.

\subsection{Anomaly detection and localization}
For anomaly detection, we further employ the F1 score and precision with a fixed recall rate of 90\%, along with the dice coefficient for localization evaluation. 
The proposed framework demonstrated superior performance in both anomaly detection and precise localization, addressing critical clinical needs for identifying subtle deviations and their spatial origins within ECG signals. For anomaly detection, the model achieved 91.2\% AUROC, 83.7\% F1 score, and 75.6\% precision (at 90\% recall) on the ECG-LT test set, significantly outperforming state-of-the-art methods such as TranAD~\cite{Tranad} (56.6\% AUROC), AnoTran ~\cite{xu2022anomaly} (60.1\% AUROC), BeatGAN~\cite{liu2022time} (65.3\% AUROC), and TSL~\cite{zheng2022task}  (82.4\% AUROC) (Tab.~\ref{tab:ad_ruijin}). Notably, it maintained robustness across varying anomaly frequencies, with AUROC scores of 91.4\%, 89.5\%, and 89.6\% for common, uncommon, and rare subsets, respectively, underscoring its effectiveness in detecting underrepresented conditions.

TSL demonstrates competitive performance in anomaly detection, achieving an AUROC of 82.4\% on all test data (Table~\ref{tab:ad_ruijin}); however, this comes at the expense of requiring substantial normal data during testing, indicating that its features lack sufficient representativeness and rely heavily on comparative analysis. Additionally, TSL’s design does not support anomaly localization, resulting in a limited AUROC score of 53.7\% and a Dice score of 56.7\%, compared to 75.6\% and 65.3\% with our method (Table~\ref{tab:ad_ruijin}). This localization constraint impairs the downstream network’s ability to focus on abnormal regions, leading to reduced diagnostic performance, particularly on rare cases (e.g., AUROC: 89.3\% \textit{vs.} 94.7\% on ECG-LT rare test data, and 76.8\% \textit{vs.} 85.4\% on Renji long-tail data, as shown in Table~\ref{tab:class_ruijin}). Furthermore, TSL’s lower sensitivity (e.g., 72.2\% \textit{vs.} 84.2\% on Renji long-tail data) underscores its challenges in detecting rare anomalies, limiting its clinical utility.

\begin{figure*}
	\centering
	\includegraphics[width=1.8\columnwidth]{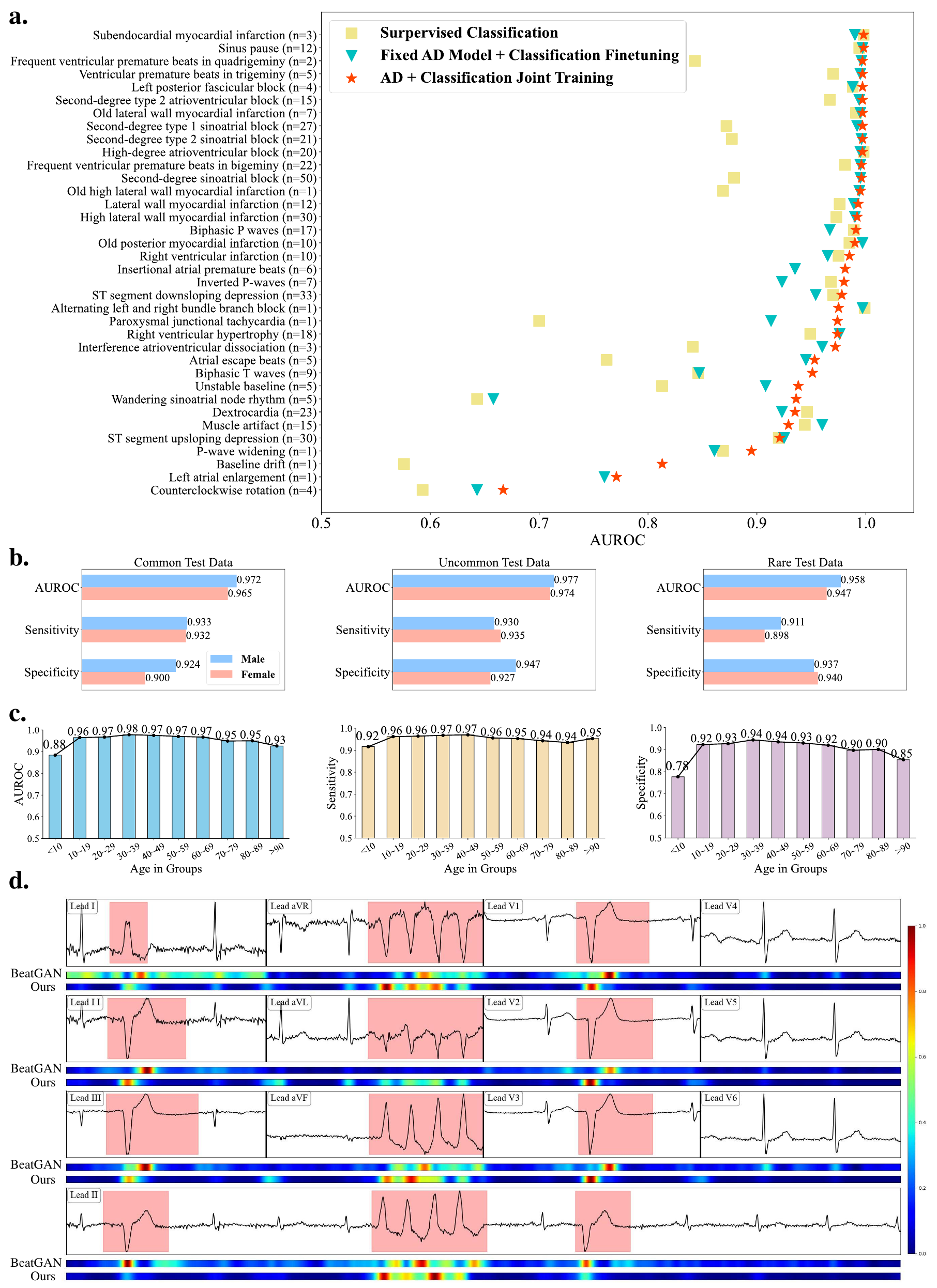}
	\caption{\textbf{Performance of ECG diagnosis.} 
(a) Diagnosis performance on tail classes. Comparison of ECG diagnosis for each type in the Rare Set, where the proposed method outperforms, particularly for anomalies with fewer samples.
(b) Diagnosis fairness across sex. The model shows consistent performance between male and female subjects, ensuring balanced accuracy.
(c) Diagnosis fairness across age. Performance is evaluated across age groups in ten-year intervals, with the model demonstrating stable and equitable results.
(d) Visualization of anomaly localization. Comparison of the proposed method (Ours) with a leading baseline (BeatGAN), against ground truth (pink boxes). Color-coded scores (0–1) indicate anomaly likelihood, with red highlighting the most probable locations.}
\label{fig:Diagnosis}
\end{figure*}

For localization, the model achieved a 76.5\% AUROC and 65.3\% Dice score on cardiologist-annotated data, surpassing the second best model, BeatGAN~\cite{liu2022time} by 11.6\% AUROC and 7.3\% Dice (Tab.~\ref{tab:ad_ruijin}).
Visual analysis of a 12-lead ECG example (Fig.~\ref{fig:Diagnosis}d) revealed precise alignment with expert annotations, particularly in identifying critical regions such as abnormal QRS complexes in Lead aVL while minimizing false positives in normal leads (e.g., Lead I). 
This capability stemmed from the multi-scale cross-attention module, which fused global signal trends with localized heartbeat features, mimicking cardiologists’ hierarchical analysis.

The model’s granular sensitivity to abrupt signal changes enabled it to highlight subtle deviations that often escape conventional methods, providing clinicians with interpretable, region-specific insights. This capability was further validated by cardiologists, who confirmed the clinical relevance of the localized anomalies in guiding diagnostic decisions.

\begin{table*}[t]
 \caption{Anomaly detection and localization results across multiple evaluation data, compared to state-of-the-art methods. The most effective method is highlighted in \textbf{bold}, and the second-best is \underline{underlined}.
 }
\centering
\scalebox{0.95}{
\begin{tabular}{cccccccc} 
\toprule
\multirow{2}{*}{Task} & Evaluation & \multirow{2}{*}{Metrics} & \multicolumn{3}{r}{State-of-the-art Methods} & & \multirow{2}{*}{Ours}\\
 & Data &  & TranAD & AnoTran & BeatGAN & TSL & \\
\Xcline{1-8}{0.3pt}
\multirow{20}{*}{\makecell[c]{Anomaly \\Detection}} & \multirow{5}{*}{\makecell{All\\ Test Data}} & AUROC  & 0.566$_{\pm0.002}$  &  0.601$_{\pm0.002}$   & 0.653$_{\pm0.002}$   & \underline{0.824$_{\pm0.002}$} & \textbf{0.912$_{\pm0.001}$}\\
& &  F1 Score & 0.667$_{\pm0.002}^{**}$  &  0.673$_{\pm0.002}^{**}$   & 0.669$_{\pm0.002}^{**}$  & \underline{0.746$_{\pm0.002}^{**}$} & \textbf{0.837$_{\pm0.002}$}\\
& & Sensitivity & 0.511$_{\pm0.002}^{**}$  &  0.579$_{\pm0.002}^{**}$   & 0.522$_{\pm0.002}^{**}$  & \underline{0.793$_{\pm0.002}^{**}$} & \textbf{0.842$_{\pm0.002}$}\\
& & Specificity & 0.514$_{\pm0.002}$  &  0.564$_{\pm0.002}^{**}$  & \underline{0.698$_{\pm0.002}^{**}$}   & 0.668$_{\pm0.002}^{**}$ & \textbf{0.830$_{\pm0.002}$}\\
& & Pre@90   & 0.501$_{\pm0.002}^{**}$  &  0.530$_{\pm0.002}^{**}$   & 0.528$_{\pm0.002}^{**}$  & \underline{0.613$_{\pm0.002}^{**}$} & \textbf{0.756$_{\pm0.002}$} \\
\Xcline{2-8}{0.3pt}
& \multirow{5}{*}{\makecell{Common\\ Test Data}} & AUROC & 0.565$_{\pm0.002}$  & 0.599$_{\pm0.002}$   & 0.650$_{\pm0.002}$   & \underline{0.820$_{\pm0.002}$} & \textbf{0.914$_{\pm0.001}$}\\
& & F1 Score  & 0.667$_{\pm0.002}^{**}$  &  0.672$_{\pm0.002}^{**}$   & 0.669$_{\pm0.002}^{**}$   & \underline{0.744$_{\pm0.002}^{**}$} & \textbf{0.839$_{\pm0.002}$}\\
& & Sensitivity  & 0.512$_{\pm0.002}^{**}$  &  0.576$_{\pm0.002}^{**}$  & 0.537$_{\pm0.002}^{**}$   & \underline{0.792$_{\pm0.002}^{**}$} & \textbf{0.852$_{\pm0.002}$}\\
& & Specificity  & 0.514$_{\pm0.002}^{**}$  &  0.564$_{\pm0.002}^{**}$   & \underline{0.679$_{\pm0.002}^{**}$}   & 0.662$_{\pm0.002}^{**}$ & \textbf{0.820$_{\pm0.002}$}\\
& & Pre@90   & 0.501$_{\pm0.002}^{**}$ &  0.529$_{\pm0.002}^{**}$  & 0.528$_{\pm0.002}^{**}$  & \underline{0.611$_{\pm0.002}^{**}$} & \textbf{0.761$_{\pm0.002}$} \\
\Xcline{2-8}{0.3pt}
& \multirow{5}{*}{\makecell{Uncommon\\ Test Data}} & AUROC  & 0.578$_{\pm0.008}$  &  0.623$_{\pm0.008}$   & 0.681$_{\pm0.007}$   & \underline{0.853$_{\pm0.006}$} & \textbf{0.895$_{\pm0.005}$}\\
& & F1 Score  & 0.667$_{\pm0.008}^{**}$ &  0.677$_{\pm0.007}^{**}$   & 0.675$_{\pm0.007}^{**}$   & \underline{0.773$_{\pm0.007}^*$} & \textbf{0.819$_{\pm0.006}$}\\
& & Sensitivity  & 0.504$_{\pm0.008}^{**}$  & 0.628$_{\pm0.008}^{**}$  & 0.552$_{\pm0.008}^{**}$   & \underline{0.753$_{\pm0.007}^*$} & \textbf{0.814$_{\pm0.006}$}\\
& & Specificity  & 0.523$_{\pm0.008}^{**}$ & 0.547$_{\pm0.008}^{**}$   & 0.719$_{\pm0.007}^{**}$   & \underline{0.805$_{\pm0.006}^*$} & \textbf{0.828$_{\pm0.006}$}\\
& & Pre@90   & 0.502$_{\pm0.008}^{**}$  &  0.536$_{\pm0.008}^{**}$   & 0.536$_{\pm0.008}^{**}$   & \underline{0.625$_{\pm0.008}^*$} & \textbf{0.700$_{\pm0.007}$}\\
\Xcline{2-8}{0.3pt}
& \multirow{5}{*}{\makecell{Rare\\ Test Data}} & AUROC  & 0.568$_{\pm0.025}$  & 0.604$_{\pm0.024}$   & 0.659$_{\pm0.024}$  & \underline{0.888$_{\pm0.016}$} & \textbf{0.896$_{\pm0.015}$}\\
& & F1 Score  & 0.667$_{\pm0.024}^{**}$  &  0.676$_{\pm0.023}^{**}$  & 0.673$_{\pm0.023}^{**}$   & \underline{0.816$_{\pm0.019}$} & \textbf{0.827$_{\pm0.019}$}\\
& & Sensitivity  & 0.513$_{\pm0.025}^{**}$ &  0.582$_{\pm0.025}^{**}$  & 0.510$_{\pm0.025}^{**}$   & \underline{0.769$_{\pm0.021}$} & \textbf{0.825$_{\pm0.019}$}\\
& & Specificity  & 0.502$_{\pm0.025}^{**}$  &  0.577$_{\pm0.025}^{**}$   & 0.746$_{\pm0.022}^{**}$  & \textbf{0.885$_{\pm0.016}$} & \underline{0.831$_{\pm0.019}$}\\
& & Pre@90   & 0.496$_{\pm0.025}^{**}$  &  0.531$_{\pm0.025}^{**}$   & 0.521$_{\pm0.025}^{**}$  & \textbf{0.693$_{\pm0.023}$} & \underline{0.692$_{\pm0.023}$} \\
\hline
\multirow{2}{*}{\makecell[c]{Anomaly \\Localization}} & \multirow{2}{*}{\makecell[c]{Localization\\ Test Set}} & AUROC  & 0.602$_{\pm0.004}$  & 0.525$_{\pm0.004}$   & \underline{0.649$_{\pm0.004}$}  & 0.537$_{\pm0.004}$ & \textbf{0.765$_{\pm0.004}$}\\
& & Dice  & 0.549$_{\pm0.004}^{**}$  &  0.511$_{\pm0.004}^{**}$  & \underline{0.580$_{\pm0.004}^{**}$}   & 0.567$_{\pm0.004}^{**}$ & \textbf{0.653$_{\pm0.004}$}\\
\bottomrule
\multicolumn{8}{l}{\small $^{*}$: Denotes a \textit{p}-value $< 0.05$ compared with the proposed Joint-train method.}\\
\multicolumn{8}{l}{\small $^{**}$: Denotes a \textit{p}-value $< 0.0001$ compared with the proposed Joint-train method.}\\
\end{tabular}
}
\label{tab:ad_ruijin}
\end{table*}

\subsection{Ablation study}
The superior performance of our framework stems from three key design elements:  (i) A novel masking and restoring technique that enables the model to learn ECG characteristics by leveraging adjacent unmasked regions; (ii) Multi-scale ECG analysis, emulating the diagnostic approach of expert cardiologists; (iii) Integration of ECG parameters with demographic factors.These elements are implemented through four core components: the Masking and Restoring (MR) module, the Multi-scale Cross-attention (MC) module, the Trend-Assisted Restoration (TAR) module, and the Attribute Prediction Module (APM). We conducted an ablation study starting with a baseline model that focuses on global ECG patterns via signal reconstruction~\cite{friedman2025unsupervised}. 
As each component was incrementally added, performance improved significantly, with the AUROC rising from 77.9\% to 91.2\% across all test data and from 73.8\% to 89.6\% for rare cases (detailed in Supplementary Table~S6). This progressive enhancement underscores the critical contribution of each component to the framework’s overall effectiveness.

The Attribute Prediction Module (APM) further enhances the model’s ability to capture associations between ECG signals and patient attributes, aiding in anomaly detection and diagnosis. The APM predicts attributes for new ECG samples, providing insights into the model’s understanding of ECG data. These attributes include patient demographics (e.g., age, gender) and physiological indicators (e.g., heart rate, PR interval). As shown in Supplementary Table~S7 and  Fig.~\ref{fig:abl_apm}, the APM performs robustly across both normal and abnormal ECGs, though accuracy is slightly higher for normal instances. For example, gender prediction accuracy drops from 86.5\% for normal ECGs to 75.9\% for abnormal ECGs, with a similar trend observed for numerical attributes, where normal ECGs exhibit smaller deviations. This aligns with the model’s pretraining on normal data, confirming its ability to effectively discern signal-attribute relationships.

\begin{figure*}[!htb]
    \centering
    \includegraphics[width = \textwidth]{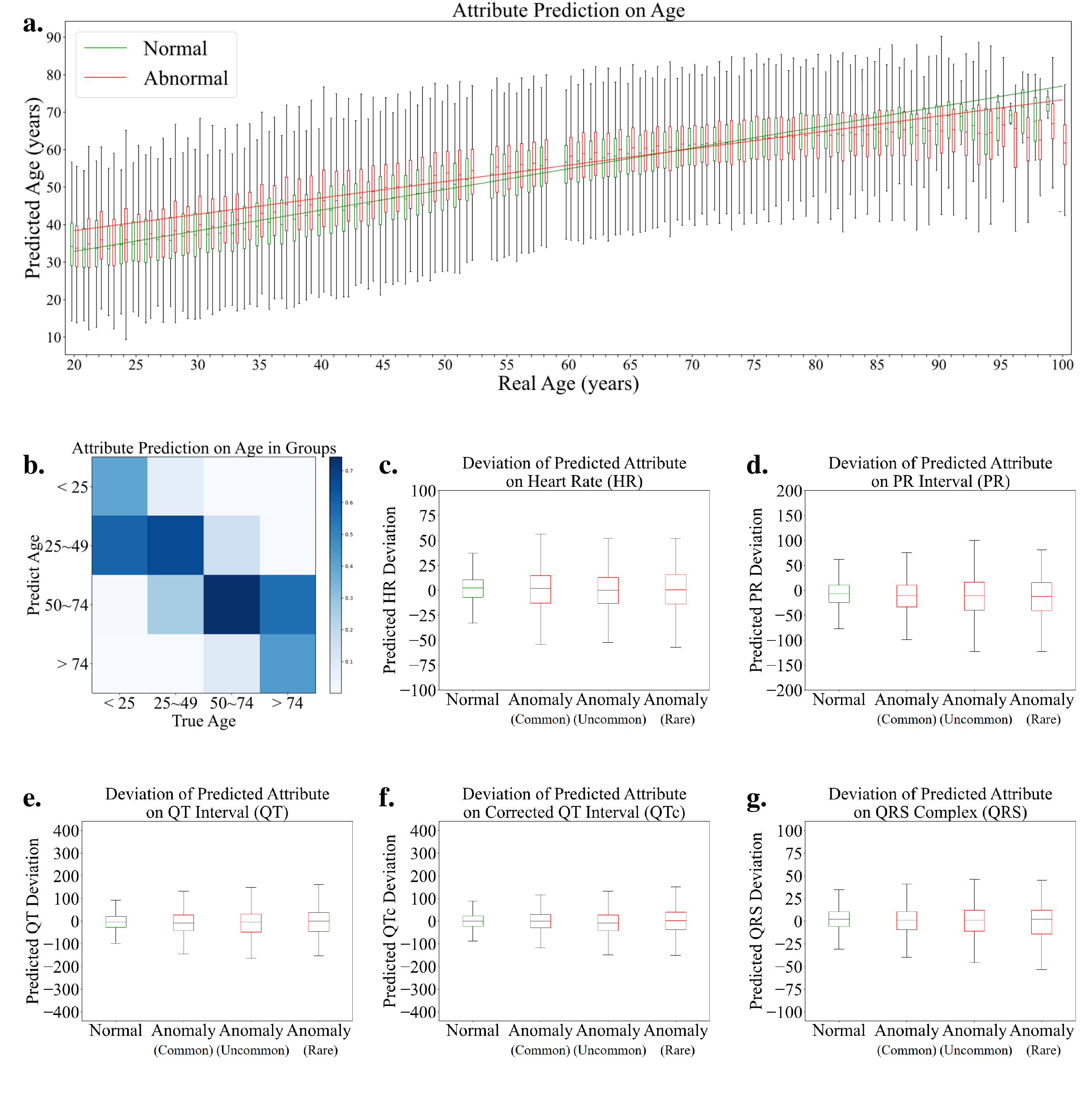}
    \caption{\textbf{Evaluation of the attribute prediction module.} \textbf{a.} Comparison of predicted ages across different age groups for normal and abnormal ECGs. \textbf{b.} Age classification accuracy for normal ECGs depicted across age ranges, showing predicted versus actual age. \textbf{c.} Heart rate prediction deviation compared to a standard reference range (y-axis) across varying anomaly rarity (x-axis). \textbf{d-g.} Analysis of deviations in PR interval, QT interval, corrected QT interval, and QRS complex predictions.}
\label{fig:abl_apm}
\end{figure*}

\subsection{Validation in a simulated clinical environment}

To assess the model’s performance in a real-world clinical setting, we deployed it in Ruijin Hospital in Shanghai to simulate a Clinical Environment in the emergency department for validation to assist cardiologists with diagnoses, without fine-tuning. The validation spanned May to July 2024, utilizing 238 distinct ECGs representing 50 unique types (see Supplementary Table~S5). ECG diagnoses in emergency department were performed under high-pressure conditions, with clinicians typically completing assessments within one minute. For comparison, each ECG was evaluated under three conditions:  
(i) Rapid diagnosis: Cardiologists provided conclusions as quickly as possible, reflecting emergency decision-making demands.  (ii) Independent diagnosis: Cardiologists diagnosed without time constts, simulating routine practice.  (iii) AI-assisted diagnosis: Cardiologists used our model’s output, offering the five most likely ECG types per case, as a reference.
All six participating cardiologists had equivalent clinical experience, ensuring unbiased comparisons across conditions. Diagnosis times were recorded discreetly to minimize bias, and an expert adjudicating cardiologist established the gold standard by reviewing all diagnoses. Performance was evaluated across three metrics: accuracy (agreement with the gold standard on key diseases), efficiency (average diagnosis time, with the model running on a single NVIDIA GTX 3090 GPU), and completeness (inclusion of detailed signal information with key disease findings).

\begin{figure*}[t]
\centering
\includegraphics[width=1.0\textwidth]{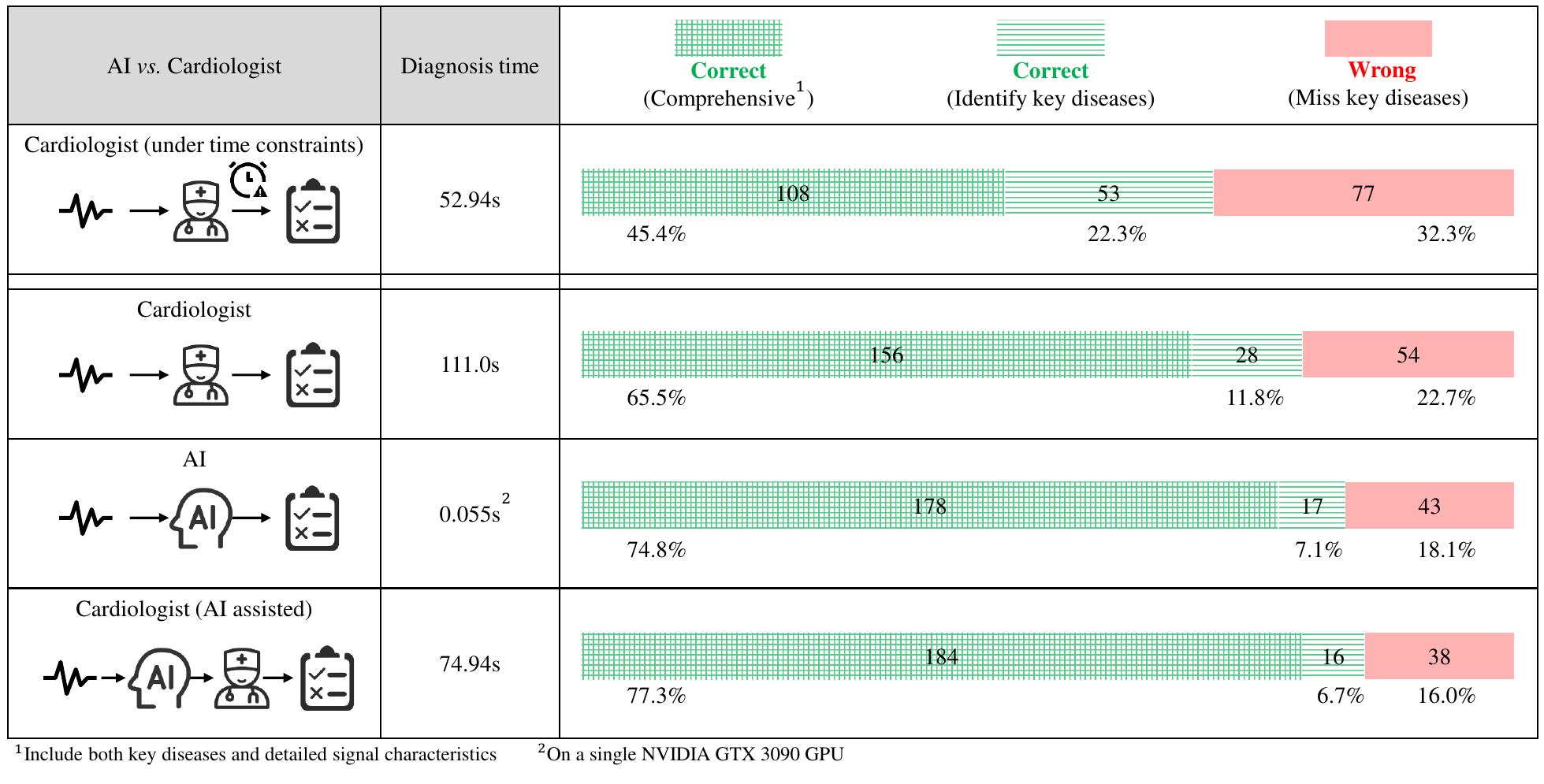}
\caption{Validation results from Ruijin Hospital’s simulated clinical environment, comparing diagnostic timing, completeness, and accuracy between cardiologists and AI-assisted diagnosis. The figure shows diagnosis time and accuracy rates for: (i) cardiologists under time constraints, (ii) cardiologists without time constraints, (iii) AI alone, and (iv) cardiologists with AI assistance. Metrics include correct identification of comprehensive diagnoses (key diseases and detailed signal characteristics), correct identification of key diseases, and frequency of missed key diseases. AI-assisted diagnosis outperformed others in speed and accuracy, notably reducing diagnostic time under time constraints.}
\label{fig:clinical_v2}
\end{figure*}

\begin{table*}[t]
 \caption{Validation results from Ruijin Hospital’s simulated clinical environment: Sensitivity and specificity analysis for common and long-tail anomalies.}
\centering
\setlength{\tabcolsep}{1.3pt}{
\begin{tabular}{cccccc} 
\toprule
Evaluation Data & Metrics & \makecell{Cardiologist \\(time constraints)} & \makecell{Cardiologist \\(no time constraints)} & AI & \makecell{Cardiologist \\(AI assisted)}\\
\hline
\multirow{2}{*}{\makecell{Common\\ data}} & Sensitivity  & 0.626$_{\pm0.165}$ & 0.747$_{\pm0.148}$ & 0.837$_{\pm0.126}$ &  0.820$_{\pm0.131}$\\
& Specificity & 0.992$_{\pm0.030}$ & 0.973$_{\pm0.055}$ & 0.916$_{\pm0.095}$ & 0.989$_{\pm0.036}$\\
\hline
\multirow{2}{*}{\makecell{Long-tail\\ data}} & Sensitivity  & 0.347$_{\pm0.226}$ & 0.469$_{\pm0.237}$ & 0.592$_{\pm0.234}$ & 0.714$_{\pm0.215}$\\
& Specificity & 0.997$_{\pm0.026}$ & 0.997$_{\pm0.026}$ & 0.986$_{\pm0.056}$ & 0.997$_{\pm0.026}$\\
\bottomrule
\end{tabular}
}
\label{tab:prospective_result}
\end{table*}

As shown in Fig.~\ref{fig:clinical_v2}, under time constraints, cardiologists demonstrated lower accuracy,  which was significantly impacted, resulting in less comprehensive conclusions that primarily focused on identifying the most clinically meaningful abnormalities (key diseases), with only 45.4\% of diagnoses being comprehensive and 32.3\% missing key diseases. In contrast, without time constraints, cardiologists achieved a comprehensive diagnosis rate of 65.5\%, missing key diseases in only 22.7\% of cases. Our AI method analyzes an ECG instance in just 0.055 seconds, approximately 1,000 times faster than the average time required for human emergency diagnosis. Beyond its speed, the AI method achieves a higher diagnostic accuracy rate of 81.9\%, outperforming the 67.7\% accuracy rate for unaided cardiologists under time constraints. When integrated into clinical practice, AI-assisted cardiologists attained an accuracy rate of 84.0\%, reflecting a 6.7\% improvement over unaided diagnoses. Additionally, diagnostic efficiency significantly improved, with an average reduction of 36 seconds in diagnosis time. The AI system also provided more detailed insights into signal patterns and rhythms, contributing to more comprehensive conclusions in 11.8\% of ECGs, particularly in recognizing subtle changes like T wave alterations and sinus tachycardia, thereby enhancing the quality of diagnostic outcomes.

In clinical diagnosis, particularly for long-tail anomalies, cardiologists under time constraints or with less experience are prone to missing diagnoses, often demonstrating high specificity (>99\%) but very low sensitivity (<50\%). 
Integrating AI into the diagnostic process significantly mitigates these errors by enhancing the detection of rare anomalies and highlighting critical signal patterns.
As illustrated in Tab.~\ref{tab:prospective_result}, our model, when used as an assistive tool, improved the sensitivity of cardiologists on long-tail data from 46.9\% to 71.4\%, while maintaining a high specificity of 99.7\%. This demonstrates the potential of AI to augment clinical decision-making, especially in challenging diagnostic scenarios.

\section{Discussion}
Our work addresses critical, long-standing gaps in AI-driven ECG diagnosis through three core innovations: (i) a self-supervised anomaly detection pretraining strategy tailored to long-tailed clinical distributions, (ii) clinically actionable anomaly localization for enhanced model transparency and trust, and (iii) robust generalizability across diverse healthcare settings and patient populations. Below, we contextualize these contributions relative to existing research and unmet clinical needs.

\textbf{Advancing long-tailed ECG diagnosis through anomaly detection pretraining.} While existing ECG diagnostic systems struggle with rare anomalies due to their reliance on class-balanced supervised training~\cite{kalmady2024development}, our anomaly detection-driven approach uniquely bridges unsupervised representation learning with practical clinical diagnostic needs. It addresses this gap through two key design elements: (i) physiologically grounded pretraining that reconstructs masked ECG segments while predicting clinically validated attributes (QT intervals, demographic characteristics), and (ii) hierarchical feature fusion that integrates global rhythm patterns and localized heartbeat-level details. This dual focus explains why our framework reduces the common-to-rare performance gap by 73\% relative to fully supervised baselines, narrowing the AUROC difference from 8.2\% to just 2.2\%, a substantial advance beyond prior anomaly detection methods like TSL, which primarily target coarse-grained binary abnormality detection. Crucially, our attribute prediction module achieves 86.5\% sex classification accuracy on normal ECGs (Supplementary Table~S7), demonstrating its capacity to model well-documented ECG-demographic relationships~\cite{moss2010gender,carbone2020gender,simonson1972effect,macfarlane2018influence}, a capability largely overlooked in conventional self-supervised pretraining frameworks.

\textbf{Anomaly localization provides clinical-grade interpretability.} Beyond diagnostic accuracy, our method’s ability to precisely localize anomalies (76.5\% AUROC) delivers actionable clinical insights by highlighting diagnostically critical regions, such as ST-segment deviations (Fig.~\ref{fig:Diagnosis}d). This transparency aligns with recent FDA guidelines for explainable AI in medical devices~\cite{FDAGuideline}, enabling clinicians to validate model reasoning against their clinical expertise, a key feature lacking in black-box deep learning alternatives. By pinpointing anomalous regions at the signal-point level, these localization outputs provide cardiologists with targeted, interpretable cues to guide their review, rather than requiring them to reverse-engineer the model’s decision-making process.

\textbf{Anomaly detection pretraining improves multi-center generalizability.} Another critical advantage of our framework lies in its robust performance across geographically and clinically diverse institutions. When evaluated on the PTB-XL and Renji datasets, each with distinct patient demographics, recording hardware, and clinical annotation protocols, our anomaly detection pretrained model outperforms baseline methods by 1.8–8.1\% in AUROC (Table~\ref{tab:class_ruijin}). This consistent performance, observed despite substantial cross-site variations, reinforces that prioritizing the modeling of normal cardiac physiology, rather than overfitting to site-specific anomaly patterns, mitigates domain shift and institutional biases. These findings align with prior work~\cite{hendrycks2019using,yousef2023no,cao2023anomaly} demonstrating that self-supervised anomaly detection frameworks are more resistant to out-of-distribution data shifts than purely supervised alternatives.

\textbf{Fairness of the clinical diagnosis models.} Ensuring unbiased clinical deployment requires not only strong overall diagnostic accuracy, but also consistent performance across key demographic subgroups, most notably age and sex. As shown in Fig.~\ref{fig:Diagnosis}b and Fig.~\ref{fig:Diagnosis}c, our analysis demonstrates comparable diagnostic efficacy between male and female patients, as well as across age groups from 10 to 90 years. We do, however, observe a modest reduction in diagnostic performance in patients under 10 years of age and those over 90, relative to other age groups. This disparity likely arises from unique physiological characteristics and atypical disease presentations at these age extremes, which are underrepresented in the training dataset. These findings highlight the importance of explicitly accounting for demographic factors to enhance both diagnostic accuracy and equity in clinical practice. Further research is needed to characterize the mechanisms driving these age-related performance differences, and to develop targeted calibration strategies to ensure equitable performance across the full age spectrum, advancing equitable care for all patient populations.

\textbf{Clinical translation.} Our trial provides robust evidence of real-world clinical utility, demonstrating that AI-assisted interpretation improves cardiologists’ sensitivity to rare anomalies by 24.5\% without compromising overall specificity (99.7\%). This advance addresses a longstanding limitation of conventional clinical decision support systems, which typically prioritize performance on common conditions at the expense of rare, life-threatening arrhythmias. Equally impactful is the 32.5\% reduction in ECG interpretation time, an advance that directly addresses workflow bottlenecks in emergency departments, where rapid, accurate assessments are critical amid the growing global burden of cardiovascular disease. Notably, quantitative analyses further show that less experienced clinicians, when supported by our framework, can achieve diagnostic performance on par with, or even exceeding, that of senior cardiologists.

\textbf{Practical deployment in resource-limited settings.} 
Beyond acute care settings, our framework demonstrates broad applicability in routine primary and community care environments. For example, its ability to deliver a foundational heart rhythm analysis aligned with a cardiologist’s initial review enables non-specialist care providers to accurately detect ECG anomalies, even when using low-cost, portable recording equipment. This capability is particularly critical in settings with unevenly distributed cardiology expertise and medical resources. We further demonstrate deployment feasibility through CPU-based optimization: we leverage PyTorch’s compiler for computational graph optimization, utilize Torch Script to eliminate Python runtime overhead, and adopt the \textit{channels\_last} memory layout to improve cache efficiency on CPU architectures. Together, these optimizations reduce average inference latency from 10.0 seconds to 2.65 seconds, with a best-case latency of 1.13 seconds, on an Intel Xeon Gold 5320 CPU, enabling real-time deployment on standard clinical hardware without specialized GPU acceleration. In China and other regions undergoing large-scale healthcare system reform, our solution can help bridge diagnostic gaps in community clinics, supporting earlier detection and proactive management of cardiovascular disease across diverse care settings.

\section{Summary}
In summary, our study demonstrates that integrating anomaly detection pretraining with demographic-aware feature learning can substantially improve both the accuracy and equity of rare cardiac anomaly diagnosis. Beyond technical performance gains, our findings highlight how equity-centered AI frameworks can be seamlessly embedded into routine clinical workflows to mitigate longstanding diagnostic disparities, improve care efficiency, and enable the delivery of more timely, consistent cardiac care for diverse patient populations.



\section{Ethics statement}
The study protocol was approved by the Ruijin Hospital Ethics Committee, Shanghai Jiao Tong University
School of Medicine (reference number: 2024-220), and conducted in accordance with the Declaration of
Helsinki~\cite{world2013world}. All data were anonymized to protect patient privacy.

\printcredits

\bibliographystyle{cas-model2-names}

\bibliography{cas-refs}

\appendix

\renewcommand{\thetable}{S\arabic{table}}
\renewcommand{\thefigure}{S\arabic{figure}}
\setcounter{figure}{0}
\setcounter{table}{0}

\clearpage

\begin{table*}[!htb]
\caption{Detailed information about common data on our general hospital ECG-LT dataset.}
\centering
\begin{tabular}{ccccc}
\hline
Data  & Condition & \multirow{2}{*}{Cardiac Conditions} & \multirow{2}{*}{\# All data} & \multirow{2}{*}{\# Test data}\\
Type & Types & & & \\
\hline
\multirow{37}{*}{Common}  & \multirow{37}{*}{37} & Normal electrocardiogram & 539,607 & 94,304\\
& & T wave changes & 182,756 & 33,907\\
& & Sinus bradycardia & 99,275 & 17,463\\
& & ST-T segment changes & 80,334 & 11,953\\
& & Sinus tachycardia & 61,287 & 11,272\\
& & ST segment changes & 49,129 & 7,177\\
& & Mild T wave changes & 45,904 & 6,692\\
& & First-degree atrioventricular block & 35,470 & 6,774\\
& & Left ventricular high voltage & 35,104 & 5,750\\
& & Atrial fibrillation & 35,090 & 95\\
& & Complete right bundle branch block & 34,614 & 6,161\\
& & Atrial premature beat & 29,376 & 5,684\\
& & Sinus arrhythmia & 27,141 & 4,277\\
& & Ventricular premature beat & 25,765 & 4,401\\
& & Low voltage & 21,759 & 4,103\\
& & Left anterior fascicular block & 20,035 & 3,842\\
& & Mild ST segment changes & 17,542 & 3,168\\
& & ST segment saddleback elevation & 16,898 & 2,698\\
& & Incomplete right bundle branch block & 15,917 & 3,045\\
& & ST segment depression & 13,246 & 1,101\\
& & Peaked T wave & 9,979 & 2,538\\
& & Clockwise rotation & 6,786 & 1,264\\
& & Atrial flutter & 6,248 & 39\\
& & Prominent U wave & 5,788 & 1,496\\
& & Complete left bundle branch block & 5,419 & 812\\
& & Ventricular paced rhythm & 5,055 & 474\\
& & Atrial tachycardia & 5,014 & 612\\
& & Intraventricular conduction block & 4,830 & 738\\
& & Left ventricular hypertrophy & 4,423 & 731\\
& & Paroxysmal atrial tachycardia & 4,185 & 598\\
& & Inferior myocardial infarction & 3,346 & 423\\
& & Mild ST-T segment changes & 3,174 & 405\\
& & Frequent atrial premature beat & 3,008 & 424\\
& & Extensive anterior myocardial infarction & 2,988 & 485\\
& & Anteroseptal Myocardial Infarction & 2,723 & 481\\
& & Frequent ventricular premature beat & 2,604 & 360\\
& & Abnormal Q wave & 2,539 & 223\\
\hline
\end{tabular}
\label{tab:dataset_common}
\end{table*}

\begin{table*}[!htb]
\caption{Detailed information about uncommon data on our general hospital ECG-LT dataset.}
\centering
\begin{tabular}{ccccc}
\hline
Data  & Condition & \multirow{2}{*}{Cardiac Conditions} & \multirow{2}{*}{\# All data} & \multirow{2}{*}{\# Test data}\\
Type & Types & & & \\
\hline
\multirow{36}{*}{Uncommon}  & \multirow{36}{*}{36} & Prolonged QT interval & 2,486 & 410\\
& & Right axis deviation & 2,476 & 450\\
& & Old anterior myocardial infarction & 2,403 & 418\\
& & Old inferior myocardial infarction & 2,344 & 362\\
& & Left axis deviation & 1,856 & 188\\
& & Atrial paced rhythm & 1,854 & 112\\
& & Pre-excitation syndrome & 1,829 & 292\\
& & Intraventricular conduction delay & 1,740 & 269\\
& & Flat T wave & 1,701 & 128\\
& & Poor R wave progression or reversed progression & 1,642 & 281\\
& & Non-conducted atrial premature beat & 1,445 & 256\\
& & Junctional escape beat & 1,444 & 142\\
& & Insertional ventricular premature beat & 1,351 & 268\\
& & Short PR interval with normal QRS complex & 1,299 & 183\\
& & Second-degree atrioventricular block & 1,287 & 217\\
& & Couplet atrial premature beat & 1,259 & 206\\
& & Atrial bigeminy & 1,244 & 223\\
& & Paroxysmal supraventricular tachycardia & 1,107 & 5\\
& & Elevated J point & 985 & 162\\
& & Paced electrocardiogram & 972 & 1\\
& & Peaked P wave & 897 & 196\\
& & Couplet ventricular premature beat & 744 & 114\\
& & Non-paroxysmal junctional tachycardia & 743 & 81\\
& & Bifid P wave & 732 & 193\\
& & Third-degree atrioventricular block & 698 & 15\\
& & Second-degree type 1 atrioventricular block & 692 & 137\\
& & Horizontal ST segment depression & 679 & 48\\
& & Junctional premature beat & 620 & 115\\
& & Ventricular tachycardia & 542 & 52\\
& & Ventricular escape beat & 509 & 30\\
& & Anterior myocardial infarction & 498 & 38\\
& & Inverted T wave & 484 & 51\\
& & Posterior myocardial infarction & 459 & 64\\
& & Long RR interval & 441 & 3\\
& & Atrial trigeminy & 436 & 77\\
& & Paroxysmal ventricular tachycardia & 428 & 51\\
\hline
\end{tabular}
\label{tab:dataset_uncommon}
\end{table*}

\begin{table*}[p]
\caption{Detailed information about rare data on our general hospital ECG-LT dataset.}
\centering
\begin{tabular}{ccccc}
\hline
Data  & Condition & \multirow{2}{*}{Cardiac Conditions} & \multirow{2}{*}{\# All data} & \multirow{2}{*}{\# Test data}\\
Type & Types & & & \\
\hline
\multirow{43}{*}{Rare}   & \multirow{43}{*}{43} & High lateral myocardial infarction & 394 & 30\\
& & Upsloping ST segment depression & 368 & 30\\
& & Downsloping ST segment depression & 311 & 33\\
& & Second-degree sinoatrial block & 257 & 50\\
& & High-degree atrioventricular block & 247 & 20\\
& & Right ventricular myocardial infarction & 213 & 10\\
& & Dextrocardia & 167 & 23\\
& & Right ventricular hypertrophy & 164 & 18\\
& & Ventricular bigeminy & 160 & 22\\
& & Second-degree type 2 sinoatrial block & 117 & 21\\
& & Second-degree type 1 sinoatrial block & 117 & 27\\
& & Atrial tachycardia with variable conduction & 115 & 0\\
& & Sinus arrest & 109 & 12\\
& & Biphasic P wave & 106 & 17\\
& & Lateral myocardial infarction & 101 & 12\\
& & Old lateral myocardial infarction & 86 & 7\\
& & Second-degree type 2 atrioventricular block & 85 & 15\\
& & Left posterior fascicular block & 80 & 4\\
& & Biphasic T wave & 65 & 9\\
& & Muscle artifact & 61 & 15\\
& & Old posterior myocardial infarction & 61 & 10\\
& & Inverted P wave & 55 & 7\\
& & Lead detachment & 51 & 0\\
& & Sinus node wandering rhythm & 48 & 5\\
& & Unstable baseline & 47 & 5\\
& & Ventricular trigeminy & 46 & 5\\
& & Subendocardial myocardial infarction & 40 & 3\\
& & Old high lateral myocardial infarction & 37 & 1\\
& & Atrial escape beat & 33 & 5\\
& & Atrial arrhythmia & 30 & 0\\
& & Broadened P wave & 23 & 1\\
& & Insertional atrial premature beat & 20 & 6\\
& & Interference atrioventricular dissociation & 20 & 3\\
& & Counterclockwise rotation & 19 & 4\\
& & Left atrial hypertrophy & 19 & 1\\
& & Right atrial hypertrophy & 19 & 0\\
& & Paroxysmal junctional tachycardia & 17 & 1\\
& & Ventricular quadrigeminy & 13 & 2\\
& & Ventricular fibrillation & 7 & 0\\
& & Shortened QT interval & 6 & 6\\
& & Alternating left and right bundle branch block & 6 & 1\\
& & Baseline drift & 4 & 1\\
& & Biauricular hypertrophy & 1 & 0\\
\hline
\end{tabular}
\label{tab:dataset_rare}
\end{table*}

\begin{table*}[p]
\caption{Detailed information about Renji dataset.}
\centering
\begin{tabular}{ccccc}
\hline
\multirow{2}{*}{Dataset}  & Condition & \multirow{2}{*}{Cardiac Conditions} & \multirow{2}{*}{\# Test data}\\
& Types & & \\
\hline
\multirow{26}{*}{Renji}   & \multirow{26}{*}{26} & Paced electrocardiogram & 398\\
& & Myocardial infarction & 223\\
& & Complete left bundle branch block & 214\\
& & Complete right bundle branch block & 209\\
& & Atrial fibrillation & 144\\
& & Ventricular premature beat & 198\\
& & Atrial premature beat & 104\\
& & Sinus tachycardia & 92\\
& & Sinus bradycardia & 62\\
& & First-degree atrioventricular block & 57\\
& & Incomplete right bundle branch block & 53\\
& & Left ventricular high voltage & 29\\
& & Atrial flutter & 21\\
& & Elevated J point & 9\\
& & Clockwise rotation & 13\\
& & Second-degree atrioventricular block & 1\\
& & Left anterior fascicular block & 11\\
& & Flat T wave & 20\\
& & Low voltage & 29\\
& & Atrioventricular block & 70\\
& & Third-degree atrioventricular block & 5\\
& & Prolonged QT interval & 2\\
& & Poor R wave progression or reversed progression & 2\\
& & Intraventricular conduction block & 2\\
& & ventricular tachycardia & 1\\
& & Second-degree type 2 atrioventricular block & 1\\
\hline
\end{tabular}
\label{tab:dataset_external}
\end{table*}

\begin{table*}[!htb]
\caption{Detailed information about data in the simulated clinical validation.}
\centering
\begin{tabular}{ccccc} 
\hline
Data  & Condition & \multirow{2}{*}{Cardiac Conditions} & \multirow{2}{*}{\#Data} & \multirow{2}{*}{Rarity}\\
Source & Types & &  & \\
\hline
   & \multirow{50}{*}{50} & Normal electrocardiogram & 40 & Common\\
& & Sinus tachycardia & 35 & Common\\
& & Atrial fibrillation & 26 & Common\\
& & ST-T segment changes & 26 & Common\\
& & T wave changes & 24 & Common\\
& & Sinus bradycardia & 21 & Common\\
& & Complete right bundle branch block & 18 & Common\\
& & Low voltage & 18 & Common\\
& & Left ventricular high voltage & 15 & Common\\
& & Atrial flutter & 15 & Common\\
& & Atrial premature beat & 12 & Common\\
& & ST segment changes & 12 & Common\\
& & Ventricular premature beat & 11 & Common\\
& & First-degree atrioventricular block & 11 & Common\\
& & Left anterior fascicular block & 10 & Common\\
& & Paroxysmal supraventricular tachycardia & 9 & Uncommon\\
& & Peaked T wave & 9 & Common\\
& & Sinus arrhythmia & 9 & Common\\
& & Prominent U wave & 9 & Common\\
& & Intraventricular conduction block & 9 & Common\\
& & Mild T wave changes & 9 & Common\\
& & Ventricular paced rhythm & 7 & Common\\
& & ST segment saddleback elevation & 7 & Common\\
& & Incomplete right bundle branch block & 6 & Common\\
Emergency & & Mild ST segment changes & 6 & Common\\
department & & Extensive anterior myocardial infarction & 5 & Common\\
& & Complete left bundle branch block & 5 & Common\\
& & Atrial tachycardia & 5 & Common\\
& & Ventricular escape beat & 5 & Uncommon\\
& & Old anterior myocardial infarction & 4 & Uncommon\\
& & Intraventricular conduction delay & 4 & Uncommon\\
& & Poor R wave progression or reversed progression & 4 & Uncommon\\
& & Frequent ventricular premature beat & 4 & Common\\
& & Frequent atrial premature beat & 4 & Common\\
& & Prolonged QT interval & 4 & Uncommon\\
& & Paced electrocardiogram & 4 & Uncommon\\
& & Couplet atrial premature beat & 3 & Uncommon\\
& & Left ventricular hypertrophy & 3 & Common\\
& & Old inferior myocardial infarction & 3 & Uncommon\\
& & Paroxysmal atrial tachycardia & 3 & Common\\
& & Junctional escape beat & 3 & Uncommon\\
& & Short PR interval with normal QRS complexs & 2 & Uncommon\\
& & ST segment depression & 2 & Common\\
& & Clockwise rotation & 2 & Common\\
& & Junctional premature beat & 1 & Uncommon\\
& & Non-conducted atrial premature beat & 1 & Uncommon\\
& & Elevated J point & 1 & Uncommon\\
& & Flat T wave & 1 & Uncommon\\
& & Left axis deviation & 1 & Uncommon\\
& & Pre-excitation syndrome & 1 & Uncommon\\
\hline
\end{tabular}
\label{tab:dataset_cohort}
\end{table*}

\begin{table*}[t]
 \caption{Ablation studies evaluating the impact of masking and restoring (MR), multi-scale cross-attention (MC), trend assisted restoration (TAR), and the attribute prediction module (APM) for anomaly detection on various test sets. Results are shown in the patient-level AUC. The best-performing method is in \textbf{bold}.
 }
 \label{tal:abl_ruijin}
\centering
\begin{tabular}{cccccc|c} 
\toprule
Evaluation & \multirow{2}{*}{Metrics} & \multirow{2}{*}{None} & \multirow{2}{*}{+MR} & \multirow{2}{*}{+MR+MC}  & +MR+MC & +MR+MC\\
Data&&&&&+TAR&+TAR+APM\\
\Xcline{1-7}{0.3pt}
\multirow{5}{*}{\makecell{All\\Test Data}} & AUROC  & 0.779  &  0.882   & 0.905   & 0.909 & \textbf{0.912}\\
&  F1 Score & 0.720  &  0.802   & 0.828  & 0.833 & \textbf{0.837}\\
& Sensitivity & 0.823  &  0.810  & 0.825  & 0.841 & \textbf{0.842}\\
& Specificity & 0.537  &  0.790 & \textbf{0.833}   & 0.823 & 0.830\\
 & Pre@90   & 0.585  &  0.684   & 0.737  & 0.749 & \textbf{0.756} \\
\Xcline{1-7}{0.3pt}
 \multirow{5}{*}{\makecell{Common\\Test Data}} & AUROC & 0.781  &  0.882   & 0.907   & 0.911 & \textbf{0.914}\\
 & F1 Score  & 0.722  &  0.802   & 0.830   & 0.836 & \textbf{0.839}\\
 & Sensitivity  & 0.830  &  0.811  & 0.843   & 0.843 & \textbf{0.852}\\
 & Specificity  & 0.529  &  0.790   & 0.812   & \textbf{0.826} & 0.820\\
 & Pre@90   & 0.590 &  0.687  & 0.742  & 0.756 & \textbf{0.761} \\
\Xcline{1-7}{0.3pt}
 \multirow{5}{*}{\makecell{Uncommon\\Test Data}} & AUROC  & 0.771  &  0.877   & 0.886   & 0.888 & \textbf{0.895}\\
 & F1 Score  & 0.708 &  0.804   & 0.809   & 0.811 & \textbf{0.819}\\
 & Sensitivity  & 0.759  & 0.792  & 0.788   & 0.809 & \textbf{0.814}\\
 & Specificity  & 0.617 & 0.822   & \textbf{0.840}   & 0.815 & 0.828\\
 & Pre@90   & 0.550  &  0.650   & 0.686   & 0.690 & \textbf{0.700}\\
\Xcline{1-7}{0.3pt}
 \multirow{5}{*}{\makecell{Rare\\Test Data}} & AUROC  & 0.738  & 0.884   & 0.884  & 0.883 & \textbf{0.896}\\
 & F1 Score  & 0.679  &  0.801  & 0.815   & 0.810 & \textbf{0.827}\\
 & Sensitivity  & 0.715  &  0.766  & 0.759   & 0.749 & \textbf{0.825}\\
 & Specificity  & 0.608  &  0.854   & 0.897  & \textbf{0.900} & 0.831\\
& Pre@90   & 0.525  &  0.678   & 0.661  & 0.664 & \textbf{0.692} \\
\bottomrule
\end{tabular}
\label{tab:ruijin_detail}
\end{table*}

\begin{table*}[t]
 \caption{Attribute prediction results for various attributes and data types on general hospital ECG dataset. Results are shown in accuracy for binary gender and averaged deviation for other attributes.}
\centering
\setlength{\tabcolsep}{1.3pt}{
\begin{tabular}{ccccccc} 
\hline
\multirow{2}{*}{Attribute} & \multirow{2}{*}{Reference Range} & \multirow{2}{*}{Normal} & \multicolumn{4}{c}{Abnormal} \\
& & & All & Common & Uncommon & Rare \\
\hline
Gender & 0 (male) or 1 (female)  & 86.5\% Acc               & 75.9\% Acc           &  76.2\% Acc         & 69.2\% Acc   & 70.9\% Acc \\
Age & 0 $\sim$ 100 years & $\pm$12.6 & $\pm$14.0 & $\pm$13.8 & $\pm$16.8 & $\pm$18.1\\
Heart Rate (HR) & 60 $\sim$ 100 bpm & $\pm$3.64 & $\pm$6.94 & $\pm$6.86 & $\pm$8.20 & $\pm$8.62\\
PR Interval (PR) & 120 $\sim$ 200 ms & $\pm$18.5 & $\pm$26.7 & $\pm$26.2 & $\pm$35.2 & $\pm$37.8\\
QT Interval (QT) & 320 $\sim$ 440 ms & $\pm$20.7 & $\pm$33.0 & $\pm$32.4 & $\pm$43.2 & $\pm$44.0\\
Corrected QT Interval (QTc) & 350 $\sim$ 440 ms & $\pm$21.2 & $\pm$32.7 & $\pm$32.1 & $\pm$42.3 & $\pm$46.4\\
QRS Complex (QRS) & 60 $\sim$ 110 ms & $\pm$5.74 & $\pm$9.76 & $\pm$9.62 & $\pm$12.3 & $\pm$13.6\\
\hline
\end{tabular}
}
\label{tab:abl_apvalue}
\end{table*}



\end{document}